\documentclass[lettersize,journal]{IEEEtran}
\usepackage{amsmath,amsfonts}
\usepackage{algorithmic}
\usepackage{algorithm}

\usepackage{array}
\usepackage[caption=false,font=normalsize,labelfont=sf,textfont=sf]{subfig}
\usepackage{textcomp}
\usepackage{stfloats}
\usepackage{url}
\usepackage{verbatim}
\usepackage{graphicx}
\usepackage{cite}
\usepackage{enumitem}
\usepackage{array, multirow, booktabs, makecell}

\usepackage{amsmath}
\usepackage{amssymb}

\usepackage{hyperref} 
\usepackage{listings} 
\usepackage{xcolor}   

\usepackage{tabularx}
\usepackage{adjustbox}

\lstdefinelanguage{json}{
	morekeywords={true, false, null}
}

\lstdefinestyle{jsonstyle}{
	language=json,
	basicstyle=\ttfamily\fontsize{7}{8}\selectfont,
	numbers=left,
	numberstyle=\tiny\color{gray},
	stepnumber=1,
	numbersep=5pt,
	backgroundcolor=\color{white},
	showspaces=false,
	showstringspaces=false,
	showtabs=false,
	frame=none,
	rulecolor=\color{black},
	tabsize=2,
	captionpos=b,
	breaklines=true,
	breakatwhitespace=false,
	keywordstyle=\color{blue},
	stringstyle=\color{orange},
	commentstyle=\color{green},
	morekeywords={true, false, null}
}

\usepackage{caption}
\begin{document}

\title{3D-Grounded Vision-Language Framework for Robotic Task Planning: Automated Prompt Synthesis and Supervised Reasoning}

\author{Guoqin Tang, Qingxuan Jia, Zeyuan Huang, Gang Chen, Ning Ji, Zhipeng Yao}

\markboth{}%
{}

\IEEEpubid{}

\maketitle

\begin{abstract}
Vision-language models (VLMs) have achieved remarkable success in scene understanding and perception tasks, enabling robots to plan and execute actions adaptively in dynamic environments. However, most multimodal large language models lack robust 3D scene localization capabilities, limiting their effectiveness in fine-grained robotic operations. Additionally, challenges such as low recognition accuracy, inefficiency, poor transferability, and reliability hinder their use in precision tasks. To address these limitations, we propose a novel framework that integrates a 2D prompt synthesis module by mapping 2D images to point clouds, and incorporates a small language model (SLM) for supervising VLM outputs. The 2D prompt synthesis module enables VLMs, trained on 2D images and text, to autonomously extract precise 3D spatial information without manual intervention, significantly enhancing 3D scene understanding. Meanwhile, the SLM supervises VLM outputs, mitigating hallucinations and ensuring reliable, executable robotic control code generation. Our framework eliminates the need for retraining in new environments, thereby improving cost efficiency and operational robustness. Experimental results that the proposed framework achieved a 96.0\% Task Success Rate (TSR), outperforming other methods. Ablation studies demonstrated the critical role of both the 2D prompt synthesis module and the output supervision module (which, when removed, caused a 67\% TSR drop). These findings validate the framework's effectiveness in improving 3D recognition, task planning, and robotic task execution.
\end{abstract}


\section{Introduction}

\IEEEPARstart{T}{he} integration of robotics and artificial intelligence (AI) has catalyzed significant advancements in the autonomous execution of complex tasks, marking a new era of \textit{embodied intelligence} where robots are not only capable of physical interactions but also capable of reasoning and decision-making through multi-modal data fusion and intelligent planning systems\cite{ref1-smartmanu, ref8-industry5.0}. In industrial automation, robotic systems have demonstrated exceptional efficiency, precision, cost-effectiveness, and safety, enabling diverse applications such as industrial equipment installation, precision instrument assembly, and logistics automation, forming the backbone of modern smart manufacturing processes\cite{ref5-industry,ref38-cgLLM}. However, achieving robust autonomy in complex environments presents one fundamental challenge: \textit{reasonable task planning}. 

\begin{figure}[H]
	\centering
	\includegraphics[width=0.5\textwidth]{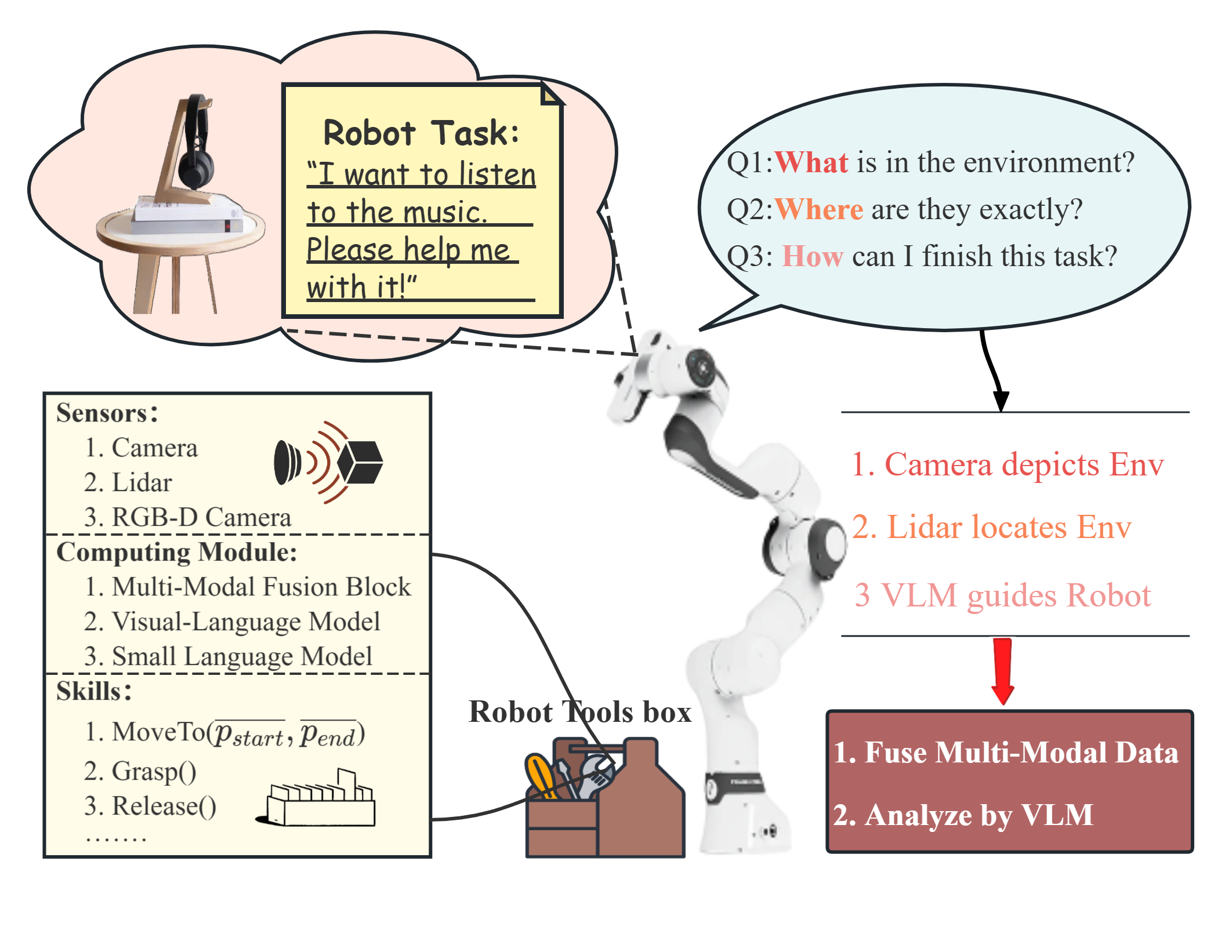}
	\caption{Overview of the robotic task execution process using a Franka robotic arm. The cloud (top left) defines the task involving objects on a table. Key challenges (top right bubble) include: \textbf{Perception}, \textbf{Localization}, and \textbf{Planning}. The proposed solution (right) integrates multimodal perception (camera \& lidar) with reasoning (VLM). The toolbox (bottom left) outlines available resources, including sensors, computing modules, and robotic skills.}
	\label{fig_0}
\end{figure}

Task planning, as stated in \cite{ref9-tampsurvey}, encompasses environmental perception, task understanding, and action planning, resulting in executable sequences of robotic actions that achieve specified objectives. Common perception techniques\cite{ref10-3ddetectcar}, ranging from image-based\cite{ref6-cvinmanu} and point cloud-based\cite{ref7-3dinmanu} methods to multimodal fusion\cite{ref11-dlfusion}, provide essential environmental modeling and serve as key constraints for subsequent planning steps. However, traditional approaches, such as rule-based methods and state-transition strategies, rely heavily on expert knowledge, predefined robot states, and fixed interaction rules\cite{ref9-tampsurvey}. This dependency limits their adaptability and scalability in complex, uncertain environments, prompting researchers to explore alternative strategies\cite{ref39-cghandoverDirection}.

\IEEEpubidadjcol

\begin{figure*}[ht]
	\centering
	\includegraphics[width=\textwidth,trim={10pt 0pt 100pt 20pt},clip]{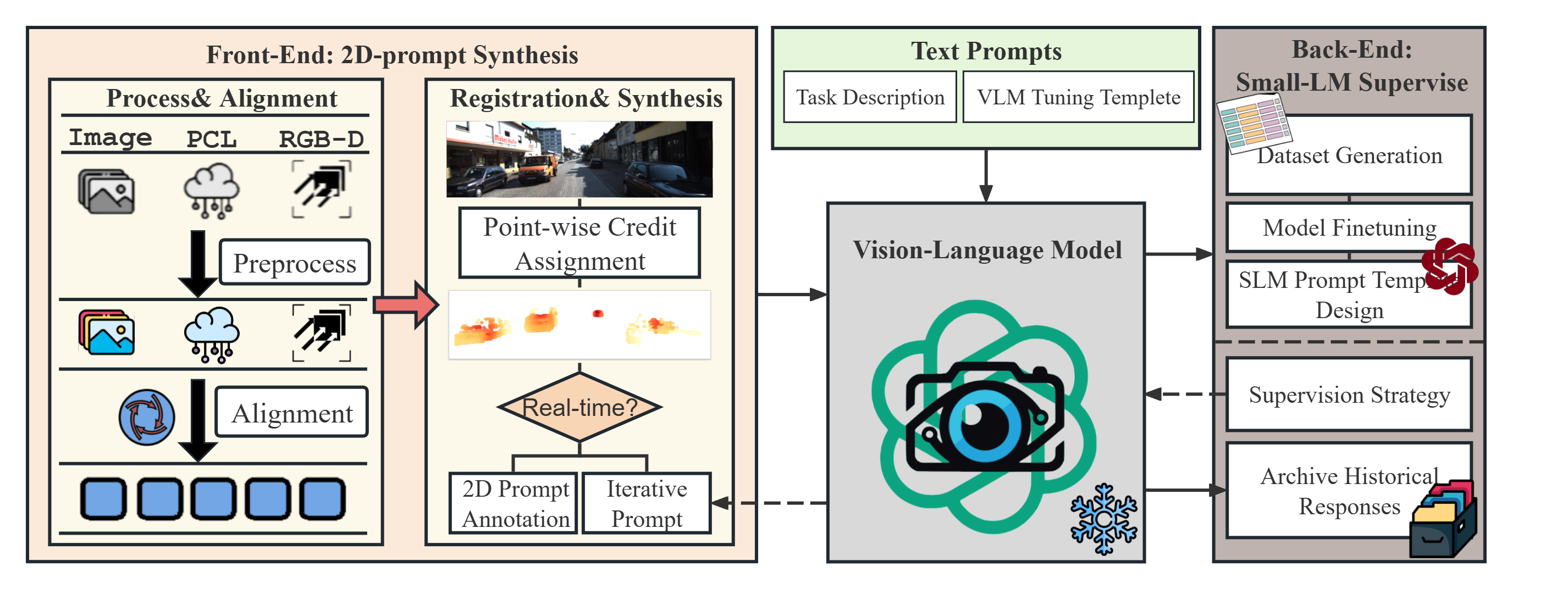}
	\caption{
 \textbf{The overall architecture of the proposed framework.} The framework consists of three main components: \textbf{2D Prompt Synthesis Module (orange)}, including \textit{Process \& Alignment} (light yellow) for multimodal data preprocessing and alignment, and \textit{Registration \& Synthesis} (light yellow) for credit-based prompt generation. A \textbf{red arrow} indicates data flow between these submodules. The \textbf{Frozen Vision-Language Model (VLM, gray)} serves as the reasoning core, receiving inputs from the \textbf{2D Prompt Synthesis Module} and \textbf{Text Prompts}. A \textbf{dashed arrow} represents iterative refinement with the \textit{iterative prompt algorithm} in \textit{Registration \& Synthesis}. The \textbf{Back-End Small Language Model (SLM) Supervision (brown)} validates and refines outputs via a \textbf{solid arrow}, with a \textbf{dashed arrow} enabling feedback correction to the VLM. Final validated outputs are archived in the \textbf{Archive Historical Responses} submodule.
	}
	\label{fig_1}
\end{figure*}

Building on these explorations, recent work has investigated multimodal fusion \cite{ref12-mmfx,ref13-RGBDconstraintMapping} for richer environmental representation and large-scale models \cite{ref14-LLM-BT,ref15-LLM3DA,ref16-CLIPFO3D} for more flexible, data-driven task planning. In these frameworks, perception is not an isolated step but an integral component of the planning pipeline, providing critical spatial and semantic cues that guide subsequent reasoning. Yet, despite these efforts, significant challenges persist in bridging the gap between high-level commands and actionable robot behaviors. For example, consider a task where a robotic arm is required to pick up a headphone from a cluttered desk and hang it on a designated stand. This scenario highlights three critical requirements. First, the system must accurately perceive the spatial information of the target object, such as its position and orientation. In complex environments, limitations in sensor resolution or perception algorithms often result in incomplete or inaccurate 3D spatial data, which directly impacts the feasibility and precision of task planning. Second, the system needs to infer geometric constraints and semantic relationships between objects, especially in multi-object interaction scenarios. For instance, it must deduce the optimal grasping point (e.g., the headphone bridge) or align objects (e.g., the headphone and the stand) based on the task requirements. Finally, the dynamic nature of the environment further complicates task execution, as unexpected changes can disrupt the scene structure. Without real-time supervision and feedback mechanisms, the system cannot adapt its plans to handle these changes, resulting in task failure. These challenges illustrate significant gaps in existing methods regarding perception, reasoning, and dynamic adaptability.

Some approaches\cite{ref17-drivegpt, ref18-progprompt} attempt to mitigate these issues by incorporating image-based cues, but they often fail to capture critical spatial relationships—such as proximity or alignment—especially in cluttered or dynamically changing scenes. Although \textit{end-to-end multimodal models} (E2E-MM)\cite{ref19-3dllm,ref20-llmi3d,ref21-rdt1b} improve perceptual accuracy by integrating multiple sensory inputs, they remain resource-intensive and difficult to interpret. Meanwhile, \textit{prompt-based modular frameworks}\cite{ref22-databaseprompt,ref23-semanticAb,ref24-MoPE}, which combine textual and visual inputs more flexibly, still struggle to fully align these modalities, limiting their spatial understanding. Moreover, the lack of robust supervision and iterative feedback loops prevents current methods from refining their plans as conditions evolve. Collectively, these challenges underscore the need for a more interpretable, spatially-aware, and feedback-driven framework that seamlessly integrates perception into the task planning process.

To address these persistent challenges, we propose a novel \textit{\textbf{plug-and-play framework}} designed to integrate multimodal data from different sensors and supervise robotic task execution in dynamic environments. By leveraging existing large-scale models without requiring extensive retraining, this approach offers a cost-effective and scalable solution. The framework consists of three key modules: (1) a \textbf{2D prompt synthesis module} at the front, which fuses depth information into 2D images to generate precise 3D spatial coordinates, enabling accurate perception of target objects; (2) a \textbf{frozen Vision-Language Model (VLM)} at the core, which leverages pre-trained visual and textual reasoning capabilities to infer geometric constraints and semantic relationships, supporting complex task planning; and (3) a \textbf{supervision module} at the back, which employs a fine-tuned small language model (SLM) to validate and refine task plans, ensuring logical consistency and dynamic adaptability. This approach enhances spatial reasoning, mitigates LLM hallucinations caused by 3D information ambiguity, and improves task planning reliability. Experiments conducted on a Franka robotic arm performing dynamic assembly tasks demonstrated significant improvements, including a 31.93\% increase in 3D recognition accuracy, a 46.40\% enhancement in localization precision, and a 58.10\% boost in task execution success rate compared to a state-of-the-art 3D-MMLM. These results validate the framework's robustness in spatial perception and multi-step planning across diverse scenarios.

Building on these improvements, our main contributions are as follows:
\begin{itemize}
	\item \textbf{A Framework Integrating 3D Perception and SLM Supervision:} A novel framework combining 2D prompt synthesis for 3D perception with SLM supervision to enhance task planning reliability and reduce hallucinations.
	\item \textbf{Credit-based Registration-Efficient Multimodal Fusion:} A high-performance algorithm integrating 2D and 3D data using confidence-based strategies for reliable and fast task execution.
	\item \textbf{Image Prompt Synthesis Algorithms for 3D Perception:} A method enabling VLMs trained on 2D data to achieve precise 3D spatial understanding and analysis.
\end{itemize}

\section{Related Work}
Achieving robust task planning in robotics requires addressing two interrelated challenges: precise 3D spatial understanding and the effective utilization of large pre-trained models. Precise 3D perception provides the geometric and spatial foundation necessary for fine-grained operations, such as grasping and trajectory planning. However, integrating these perceptual insights into large models, such as Vision-Language Models (VLMs) and Large Language Models (LLMs), is equally critical to ensure task-relevant outputs and dynamic adaptability. This section reviews progress in these areas, focusing on multimodal perception and model adaptation strategies, and highlights their limitations.
\subsection{Multimodal Perception for Robotic Operations}

Multimodal perception enhances 3D scene understanding by combining RGB, depth, and textual inputs. Current approaches can be broadly categorized into \textbf{End-to-End Models} and \textbf{Modular Architectures}. End-to-End Models, for example "RT series"\cite{ref25-rt1,ref26-rt2} and "RobotFlamingo"\cite{ref27-flamingo}, employ transformer-based frameworks to encode sensory inputs into unified cross-modal embeddings, enabling robust multimodal reasoning. These methods excel in learning rich representations but often lack geometric fidelity, struggle with task-specific adaptation, and require costly retraining on large annotated datasets, making them less effective in dynamic environments.

Modular architectures decouple perception and reasoning processes, leveraging pre-trained VLMs to reduce computational demands. For example, CLIP-based methods extract semantic features, while auxiliary modules handle spatial reasoning \cite{ref28-clipgrasp,ref29-clipFO3D}. Another approach integrates Reinforcement Learning (RL) modules to generate desired outputs directly \cite{ref30-surveyofModularLLMRL}. Despite their flexibility, these architectures face significant challenges in multimodal data fusion, particularly in addressing sensor uncertainties and prioritizing task-relevant features in cluttered or dynamic scenarios. These limitations highlight the need for adaptive frameworks capable of integrating 3D perception with dynamic task planning.

\begin{figure}[t]
	\centering
	\includegraphics[width=0.5\textwidth]{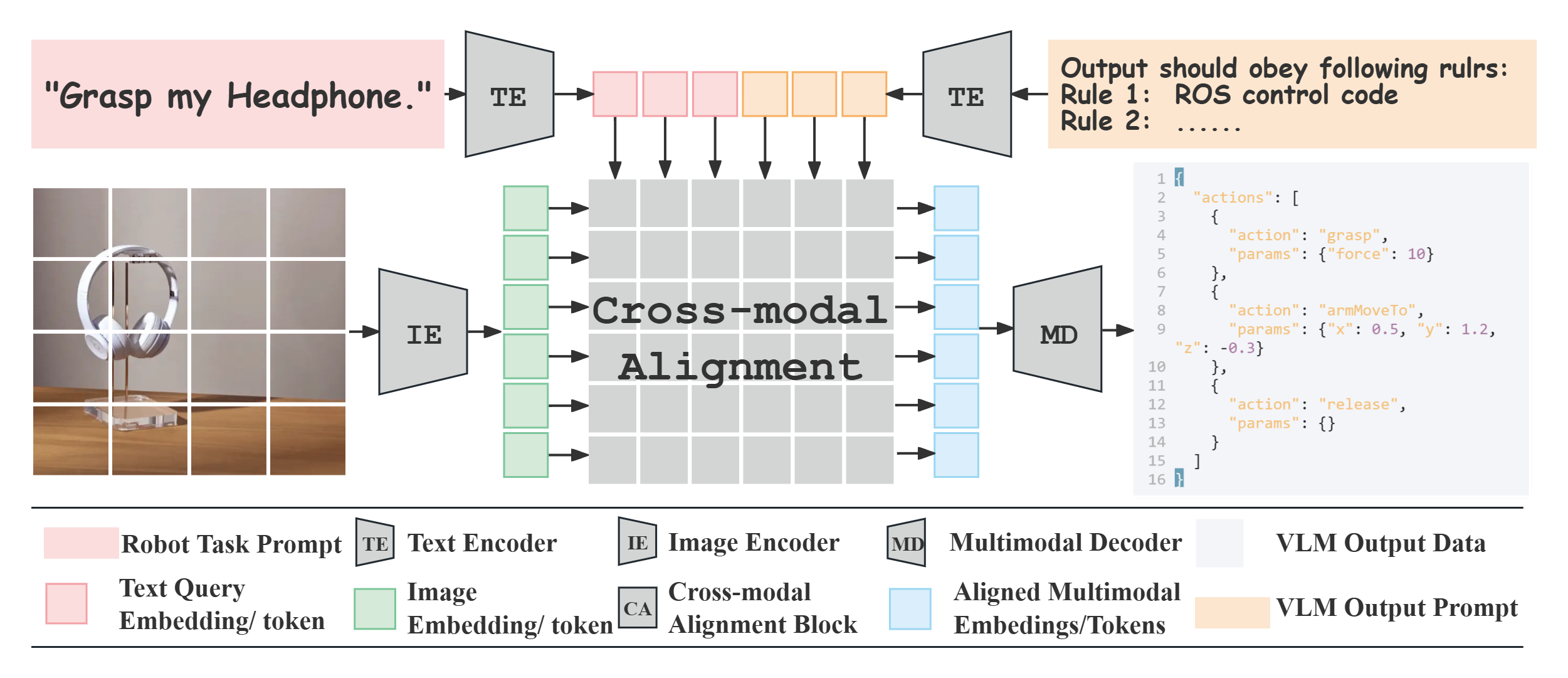}
	\caption{
		The architecture of the Vision-Language Model (VLM). Inputs include a \textbf{robot task} (top-left), \textbf{segmented image} (bottom-left), and \textbf{text template} (top-right), processed by encoders to generate feature vectors. The \textbf{cross-modal alignment module} (center) integrates text and image features, producing a unified representation. The \textbf{multimodal decoder} (right) generates executable robot control codes.
	}
	
	\label{fig_2}
\end{figure}

\subsection{Large Pre-trained Models In Robotic Tasks}

Large pre-trained models (VLMs and LLMs) have emerged as powerful tools for robotic task planning \cite{ref4-VLM-MSGM} and navigation \cite{ref2-visnav} through their multimodal reasoning capabilities \cite{ref3-autodrive}. Their integration into robotics employs two complementary approaches: \textit{prompt engineering} that steers models through input design without parameter modification, and \textit{architectural innovations} involving structural adaptations or domain-specific retraining. 

Prompt engineering strategies, which preserve original model parameters, include textual instructions for task decomposition \cite{ref33-saycan} and visual prompts encoding spatial layouts via images \cite{ref31-Blip2,ref32-LLAVAnext}. Advanced techniques like chain-of-thought (CoT) prompting \cite{ref43-cotRobot} further enhance complex task handling by breaking operations into logical sequences. However, their reliance on textual or 2D visual inputs introduces geometric reasoning limitations, leading to failures in precision tasks such as sub-centimeter alignment. 
Architectural innovations address these limitations at the cost of flexibility: Models like XComposer \cite{ref34-xcomposer} leverage cross-modal attention mechanisms to align visual-textual embeddings obtaining geometric grounding ability. Mini-CPM \cite{ref35-minicpm}, a new cross-modal attention work, optimizes computational efficiency through layer pruning. This type of mechanism(multimodal pretrained method) underpin diverse applications: Well-designed domain-specific VLMs guide navigation in large-scale environments \cite{ref40-VLMap,ref41-ViNG}, while multimodal LLMs like PALM-E \cite{ref42-Palm-E} enable robotic arm control through embodied reasoning. Although performing well in specific domain, structure-modified models suffer from resource-intensive retraining for new environment. 

These challenges highlight three key unresolved issues: the semantic-geometric divide in complex 3D tasks, the computational burden of architectural adaptations, and the absence of automatic geometric grounding in prompt-based approaches.

\section{Methodology}

This paper introduces a modular framework (Fig.~\ref{fig_1}) to address the limitations of Vision-Language Models (VLMs) in fine-grained robotic task planning. The proposed framework enhances spatial reasoning, logical consistency, and adaptability by integrating three complementary modules, enabling precise and reliable execution of complex robotic tasks.

\subsection{VLM-Based Robotic Task Planning Framework}

The proposed framework leverages a \textbf{frozen Vision-Language Model (VLM)} pre-trained on large-scale datasets, as shown in Fig.~\ref{fig_2}. While traditional VLMs excel in image-text reasoning, they lack precise 3D spatial understanding and efficient multimodal integration, limiting their effectiveness in robotic task planning.

To address these limitations, the framework adopts a modular architecture consisting of the following key components:

\begin{itemize}
	\item \textbf{2D Prompt Synthesis Module:} This module processes and aligns multimodal data, ensuring accurate integration of RGB and depth information. It employs registration algorithms and iterative prompting strategies to optimize task-specific prompts for downstream reasoning tasks.
	
	\item \textbf{Frozen Vision-Language Model:} Serving as the reasoning core, the VLM integrates 2D prompts and textual commands for multimodal reasoning and task plan generation. Iterative interactions with the synthesis module enable refined and context-aware outputs.
	
	\item \textbf{Back-End Small Language Model Supervision:} This module validates and refines outputs from the VLM, ensuring logical consistency and actionable task plans. Approved outputs are archived for iterative refinement and future reference.
\end{itemize}

This modular design bridges the gap between 2D VLM capabilities and 3D robotic requirements, offering a solution that supports dynamic environments, improves spatial reasoning, and ensures robust task execution. Applications include precision assembly, dynamic navigation, and complex task execution. Detailed implementations of each module are presented in the following sections.

\begin{figure*}[ht]
	\centering
	\includegraphics[width=\textwidth,trim={40pt 40pt 40pt 50pt},clip]{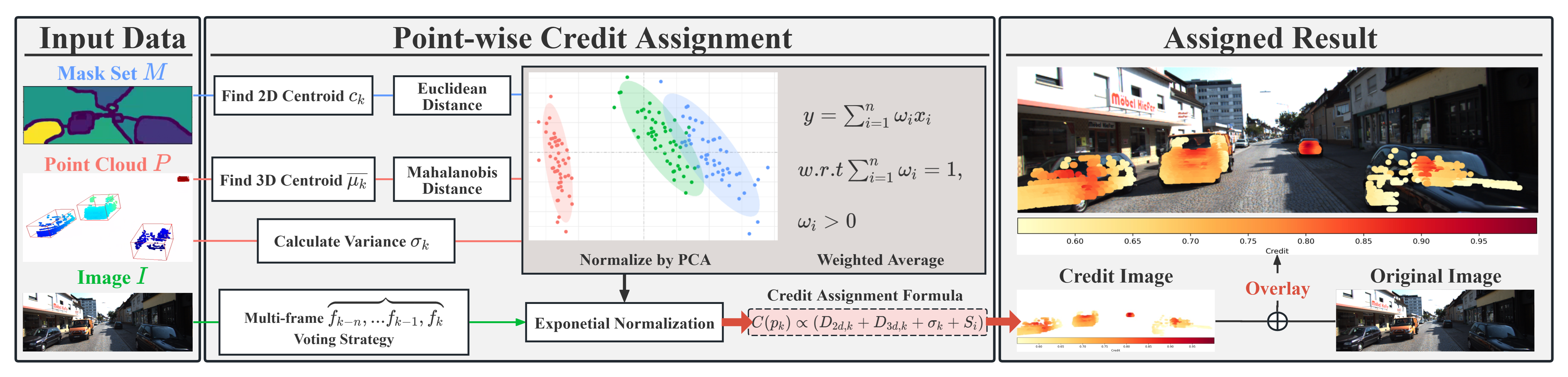}
	\caption{The figure illustrates the process of computing confidence scores using filtered point cloud data and corresponding paired image data. The central section outlines the computational steps, including mask processing, 3D point cloud analysis, and RGB image integration. The rightmost block presents the final assignment results, where confidence scores are overlaid onto the original image for visualization.}
	\label{fig_3}
\end{figure*}

\subsection{Confidence-Based Registration Strategy}
\label{confidence}

Accurate multimodal data fusion is essential for robotic perception systems, yet it remains challenging due to segmentation errors, sensor noise, and spatial misalignment. These uncertainties are especially detrimental in dynamic environments or tasks requiring high spatial accuracy. To address these issues, we propose a confidence-based registration strategy that evaluates and prioritizes spatial points based on an entropy-guided probabilistic framework. This strategy ensures that only reliable information contributes to downstream tasks, such as object localization and trajectory planning.

\subsubsection{Mathematical Definition of Confidence Score}

The confidence score \(C_i\) quantifies the reliability of each spatial point \(x_i\), integrating multiple sources of uncertainty into a unified probabilistic measure:
\begin{equation}
	C_i = \exp\left(-\sum_{n=1}^{K} \lambda_n \cdot H_n(x_i)\right),
	\label{eq:confidence_score}
\end{equation}
where \(H_n(x_i)\) represents the normalized entropy capturing uncertainty in the \(n\)-th dimension, \(\lambda_n\) is the task-specific weighting coefficient reflecting the importance of the \(n\)-th dimension, and \(K\) is the total number of uncertainty dimensions.

This formulation draws from information theory, where entropy quantifies the level of uncertainty in a probability distribution. By exponentially scaling the summed entropies, this model ensures that points with higher uncertainty have significantly lower confidence scores, prioritizing reliable points for fusion and downstream tasks.

\subsubsection{Entropy Components and Their Physical Significance}

The confidence score integrates four entropy components, each addressing distinct uncertainty sources. These components are normalized to the range \([0, 1]\) to ensure uniform scaling and compatibility across heterogeneous data modalities.

\paragraph{Spatial Consistency Entropy \(H_1(x_i)\)}

This component measures the deviation of a point \(x_i\) from the centroid of its segmentation mask \(\mathbf{c}_k\). Using the 2D Euclidean distance:
\begin{equation}
	d_{\text{2D}}(x_i) = \|\mathbf{p}_i - \mathbf{c}_k\|_2,
\end{equation}
where \(\mathbf{p}_i\) is the 2D position of \(x_i\). The normalized probability and entropy are:
\begin{equation}
	P_1(x_i) = \frac{1}{1 + d_{\text{2D}}(x_i)}, \quad H_1(x_i) = -P_1(x_i) \log(P_1(x_i)).
\end{equation}
Physically, \(H_1(x_i)\) quantifies the consistency of \(x_i\) within its segmentation context, penalizing outliers far from the centroid.

\paragraph{Geometric Consistency Entropy \(H_2(x_i)\)}

Geometric consistency evaluates the deviation of a point \(x_i\) from the local point cloud structure using the Mahalanobis distance:
\begin{equation}
	D_{\text{M}}(x_i) = (\mathbf{x}_i - \boldsymbol{\mu})^\top \mathbf{\Sigma}^{-1} (\mathbf{x}_i - \boldsymbol{\mu}),
\end{equation}
where \(\boldsymbol{\mu}\) and \(\mathbf{\Sigma}\) represent the local mean and covariance matrix. The normalized probability and entropy are:
\begin{equation}
	P_2(x_i) = \frac{D_{\text{M}}(x_i)}{D_{\text{max}}}, \quad H_2(x_i) = -P_2(x_i) \log(P_2(x_i)).
\end{equation}
This entropy component penalizes points deviating from the local geometric structure, ensuring robust point cloud consistency.

\paragraph{Depth Measurement Entropy \(H_3(x_i)\)}

Depth uncertainty is modeled using local depth variance \(\sigma_{z,i}^2\):
\begin{equation}
	P_3(x_i) = \frac{\sigma_{z,i}^2}{\sigma_{\text{max}}}, \quad H_3(x_i) = -P_3(x_i) \log(P_3(x_i)).
\end{equation}
Here, \(H_3(x_i)\) reflects the reliability of depth measurements, penalizing inconsistent or noisy depth readings.

\paragraph{Temporal Stability Entropy \(H_4(x_i)\)}

In dynamic environments, temporal stability measures the variability of a point \(x_i\) across frames:
\begin{equation}
	S_i = \frac{1}{T-1} \sum_{t=1}^{T-1} \frac{|x_i^{(t)} - x_i^{(t-1)}|}{D_{\text{max}}},
\end{equation}
where \(|x_i^{(t)} - x_i^{(t-1)}|\) represents the displacement between consecutive frames. The normalized probability and entropy are:
\begin{equation}
	P_4(x_i) = \frac{S_i}{S_i + 1}, \quad H_4(x_i) = -P_4(x_i) \log(P_4(x_i)).
\end{equation}
This entropy component penalizes points exhibiting high temporal instability, ensuring that only stable points are prioritized.

\subsubsection{Task-Driven Weight Allocation}

The weighting coefficients \(\lambda_n\) are tailored to specific tasks, prioritizing relevant uncertainty dimensions based on operational requirements:
\begin{itemize}
	\item \textbf{High spatial resolution tasks}: Emphasize \(H_1(x_i)\) and \(H_2(x_i)\) for precise spatial alignment.
	\item \textbf{Dynamic environments}: Prioritize \(H_4(x_i)\) for motion tracking and stability.
	\item \textbf{Cluttered environments}: Adjust \(\lambda_3\) to suppress noisy depth information.
\end{itemize}

For instance, in robotic navigation, temporal stability (\(H_4(x_i)\)) is critical for maintaining trajectory coherence, while precision assembly tasks emphasize spatial (\(H_1(x_i)\)) and geometric (\(H_2(x_i)\)) consistency.

\subsubsection{Integration with Downstream Tasks}

The confidence-based registration strategy ensures that high-confidence points contribute to downstream processes, such as trajectory planning and control generation. By systematically integrating multiple dimensions of uncertainty, this approach enhances the robustness and reliability of multimodal data fusion, significantly improving robotic perception and decision-making in complex environments.

\begin{figure*}[ht]
	\centering
	\includegraphics[width=\textwidth,trim={50pt 40pt 50pt 50pt},clip]{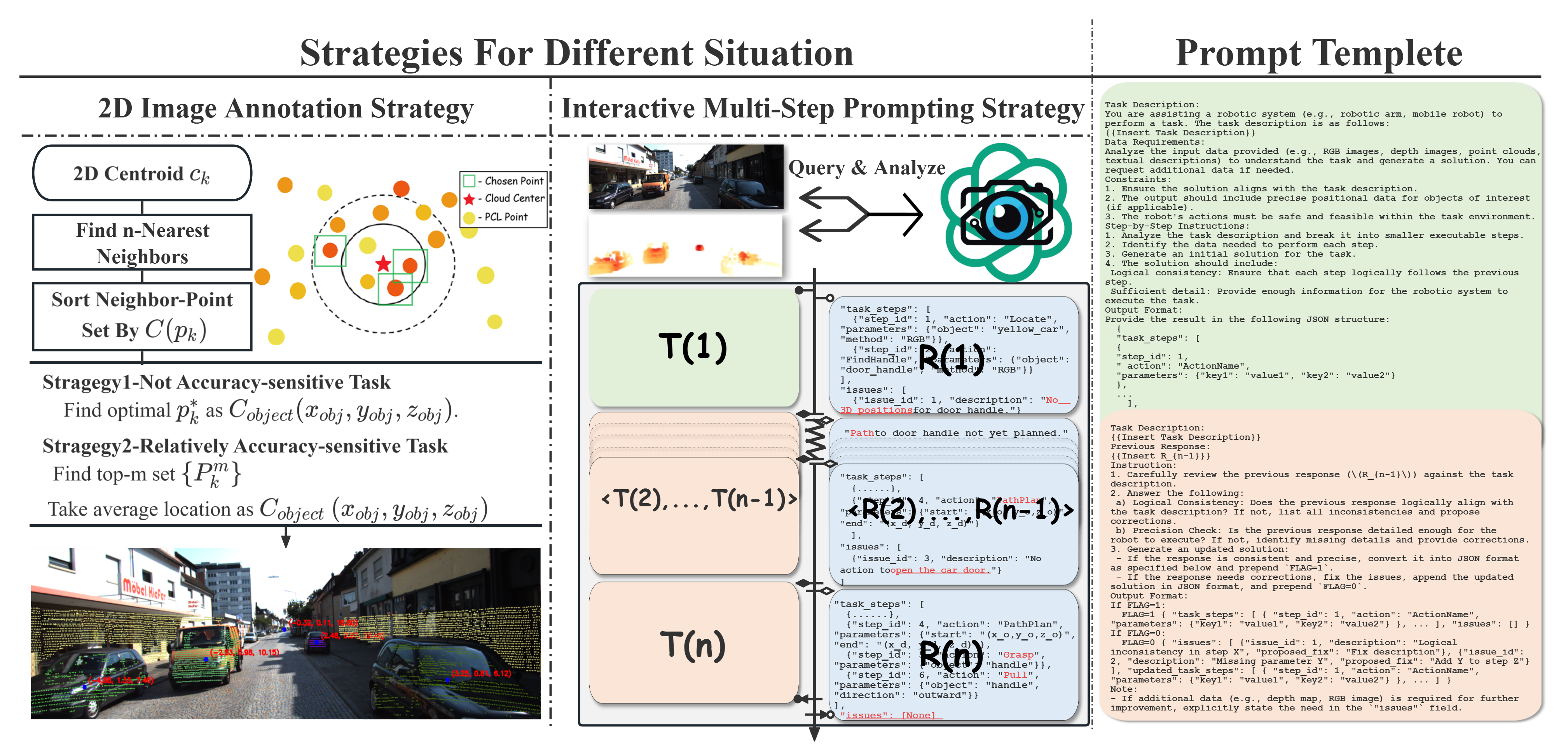}
	\caption{The figure illustrates confidence-driven strategies for task-specific prompting: The left section represents time-sensitive tasks, employing a \textbf{2D Image Annotation Strategy}, and focusing on quickly filtering and annotating key credit points; The middle section corresponds to precision-sensitive tasks, employing an \textbf{Interactive Multi-Step Prompting Strategy}, and showcasing the query and analyze interaction between the Vision-Language Model (VLM), Credit Image, and RGB Image for iterative refinement. The right section provides a Prompt Template, with the green and pink blocks aligning with their corresponding components in the middle section, offering structured guidance for logical and accurate task execution.}
	\label{fig_4}
\end{figure*}

\subsection{2D Prompt Synthesis and Textual Prompt Design}

This subsection presents a framework(fig \ref{fig_3}) that integrates 2D prompt synthesis with textual prompt design to enhance the Vision-Language Model (VLM)'s spatial reasoning capabilities. By embedding reliable 3D spatial information into 2D inputs and dynamically refining textual prompts, the framework constrains the high-dimensional space of VLM encoding, ensuring accurate outputs for robotic control tasks.

\subsubsection{Nearest Neighbor Selection Strategy}

The nearest neighbor strategy is designed for tasks requiring quick responses with moderate precision. It embeds selected 3D points into 2D inputs, reducing visual redundancy while preserving essential spatial details. For each segmentation mask $M_k$, the centroid $c_k$ in the 2D plane is computed as:
\begin{equation}
	\label{eq:centroid}
	c_k = \frac{1}{|M_k|} \sum_{\mathbf{p} \in M_k} \mathbf{p},
\end{equation}
where $|M_k|$ is the number of pixels in $M_k$, and $\mathbf{p} = (x, y)$ represents pixel coordinates.

Using the nearest neighbor (NN) algorithm, candidate 3D points $\mathcal{N}_k$ near $c_k$ are identified:
\begin{equation}
	\label{eq:nearest_neighbors}
	\mathcal{N}_k = \{\mathbf{p} \in \mathcal{P} \mid \mathbf{p} \in \text{NN}_4(c_k)\},
\end{equation}
where $\mathcal{P}$ is the point cloud. A confidence score $C(\mathbf{p})$ determines the most reliable point:
\begin{equation}
	\label{eq:most_reliable_point}
	\mathbf{p}^*_k = \arg\max_{\mathbf{p} \in \mathcal{N}_k} C(\mathbf{p}).
\end{equation}
The selected point $\mathbf{p}^*_k$ is annotated onto the 2D image, providing critical spatial cues to the VLM.

This process constrains the high-dimensional space of VLM encoding by incorporating key 3D spatial information into the image vector $\mathbf{v}_I$, which refines the probability distribution of the output vector $\mathbf{v}_O$:
\begin{equation}
	p(\mathbf{v}_O \mid \mathbf{v}_I) \rightarrow p(\mathbf{v}_O \mid \mathbf{v}_I, \mathbf{p}^*_k).
\end{equation}
The reduced uncertainty allows the decoder to generate more precise robotic control commands.

The textual prompt specifies the VLM’s role, linking the red points in the image to their corresponding 3D coordinates and instructing the model to generate robot-accessible outputs.

\subsubsection{Iterative Multi-Step Prompting Strategy}
\label{multistep strategy}
This strategy systematically refines spatial understanding and task precision through recursive visual-linguistic processing. It integrates multimodal inputs (such as images and textual descriptions) with iterative updates to the task prompt. The process consists of the following stages:

\textit{Initial Encoding and Prompt Formulation:}

The initial task prompt \( T^{(1)} \) is generated by combining the input image \( \mathcal{I} \) with the corresponding textual description \( \mathcal{D} \), along with a predefined text template \( \mathcal{T} \) that specifies the output format and interaction guidelines for the Vision-Language Model (VLM). This multimodal information is mapped into a task vector \( \mathbf{v}_T \) within a high-dimensional space \( \mathcal{V} \), as shown below:

\[
T^{(1)}: (\mathcal{I}, \mathcal{D}, \mathcal{T}) \rightarrow \mathbf{v}_T, \quad \mathbf{v}_T \in \mathcal{V}
\]

Here, \( \mathcal{I} \) is the input image, \( \mathcal{D} \) is the associated textual description (e.g., task instructions or goals), and \( \mathcal{T} \) is the predefined text template that outlines the expected output format, enabling the VLM to generate responses that include self-evaluation and check for the need for further information.

\begin{algorithm}[H]
	\caption{2D Prompt Synthesis Algorithm}
	\label{alg:prompt_synthesis}
	\textbf{Input:} Segmentation masks \(\{M_k\}\), point cloud \(\mathcal{P}\), VLM model, textual prompt templates\\
	\textbf{Output:} Enhanced 2D inputs and executable robotic commands
	\begin{algorithmic}[1]
		\STATE Initialize an empty database for storing feedback and motion primitives.
		\FOR{each segmentation mask \(M_k\)}
		\STATE Compute the centroid \(c_k\) using Equation~\ref{eq:centroid}.
		\STATE Identify the four nearest points \(\mathcal{N}_k\) in \(\mathcal{P}\) using Equation~\ref{eq:nearest_neighbors}.
		\STATE Select the most reliable 3D position \(\mathbf{p}^*_k\) using confidence scores \(C(\mathbf{p})\).
		\STATE Annotate \(\mathbf{p}^*_k\) on the 2D image near \(c_k\).
		\ENDFOR
		\STATE Initialize prompt sequence \(T^{(1)}\) for interactive refinement.
		\FOR{each iteration \(n\)}
		\STATE Generate output \(R^{(n)}\) using Equation~\ref{eq:iterative_prompting}.
		\STATE Update \(T^{(n+1)}\) based on \(R^{(n)}\).
		\STATE \textbf{IF} convergence criteria met \textbf{THEN} break.
		\ENDFOR
		\STATE Populate textual templates with task-specific parameters, leveraging database-stored motion primitives and constraints.
	\end{algorithmic}
\end{algorithm}

\textit{Iterative Refinement Mechanism:}

For each iteration \( n \), capture new image if required. Otherwise, use the previous image data.

The task prompt \( T^{(n)} \) is updated based on the image data \( \mathcal{I}^{(n)} \), the task description \( \mathcal{D} \), and the previous output \( T^{(n-1)} \):
\begin{equation}
	T^{(n)} = \text{Update}(T^{(n-1)}, \mathcal{I}^{(n)}, \mathcal{D})
\end{equation}

The Vision-Language Model (VLM) generates a response \( R^{(n)} \), which includes the region of interest (ROI) coordinates, based on the updated prompt \( T^{(n)} \) and image data \( \mathcal{I}^{(n)} \):
\begin{equation}
	\label{eq:iterative_prompting}
	R^{(n)} = \text{VLM}(\mathcal{I}^{(n)}, T^{(n)}, \mathcal{V})
\end{equation}

The coordinates of the ROI are directly extracted from \( R^{(n)} \), which also contains information about the target region in the image:
\begin{equation}
	\text{ROI}^{(n)} = \text{ExtractROI}(R^{(n)})
\end{equation}

The region of interest \( \text{ROI}^{(n)} \) is then mapped with the depth information:
\begin{equation}
	\text{Depth}_{\text{optimal}}^{(n)} = \underset{\mathbf{p} \in \text{NN}(\text{ROI}^{(n)})}{\operatorname{arg\,max}}\, C(\mathbf{p})
\end{equation}

This process iteratively refines the VLM’s output, using regions of interest to guide the robot to precise interaction points based on the confidence of depth values.

\textit{Convergence Criteria:}
\begin{equation}
	\text{Stop Condition}: |\Delta \text{ROI}^{(n)}| < \epsilon \ \text{and} \ FLAG = 1
\end{equation}

Terminate the iteration when the change in ROI falls below the threshold \( \epsilon \) and the task is deemed complete by the VLM with \( FLAG = 1 \), indicating sufficient precision has been achieved.

\subsubsection{Integration and Complementarity}

The nearest neighbor strategy emphasizes efficiency, making it suitable for tasks with strict time constraints, while the iterative strategy ensures precision, excelling in complex scenarios. By combining both approaches, we achieve a robust framework that enhances spatial understanding and interaction precision across diverse robotic applications. The integration of 2D prompt synthesis and textual design ensures that the high-dimensional encoding space remains well-constrained, enabling reliable task execution.

\begin{figure*}[ht]
	\centering
	\includegraphics[width=0.8\textwidth, trim={50px 40px 40px 40px}, clip]{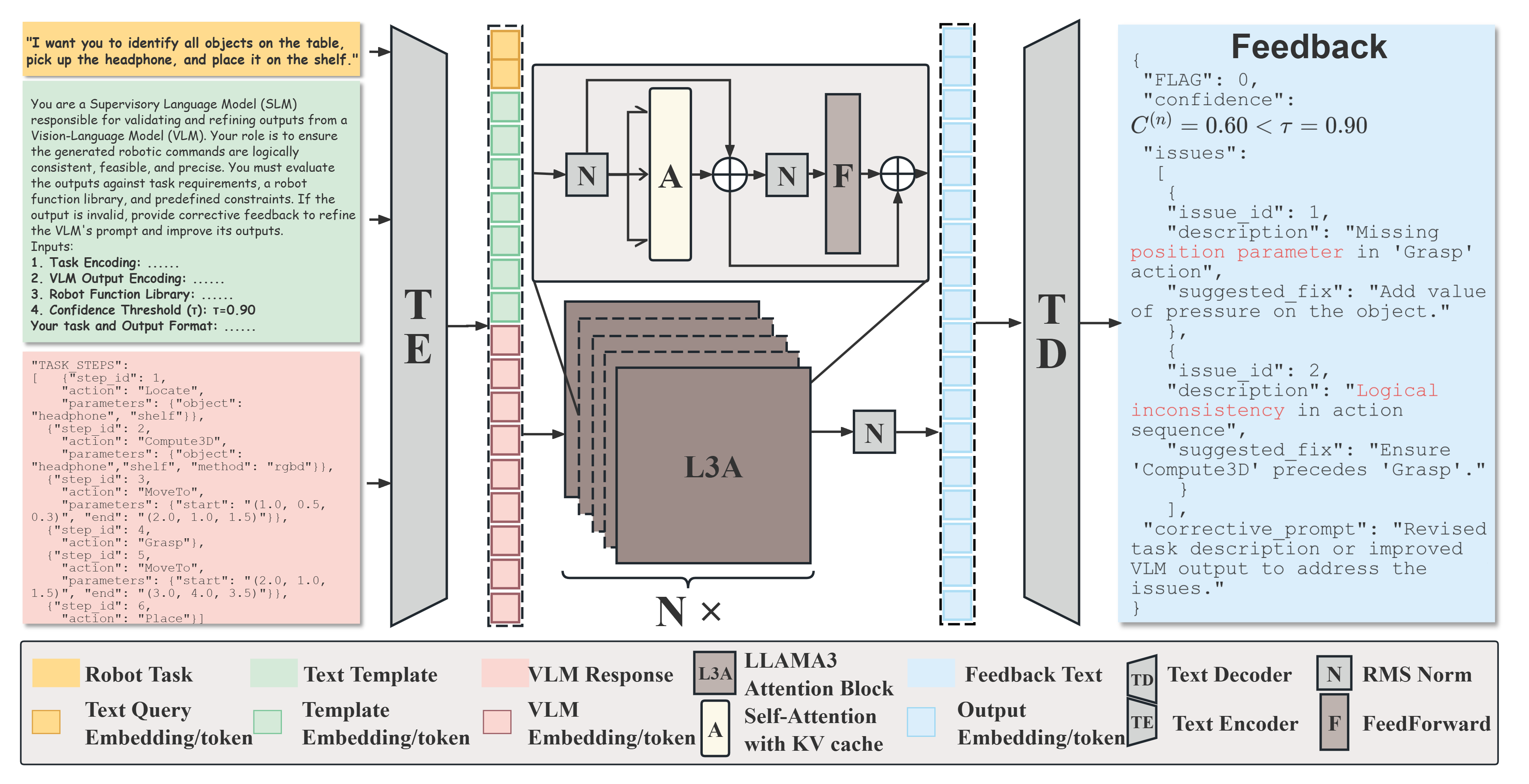}
	\caption{
		The architecture of the Small Language Model (SLM) supervision module. Inputs (left) are processed by the \textbf{text encoder} to generate embeddings, which pass through the \textbf{LLAMA3 attention block} (center) for reasoning and alignment. The outputs are decoded into actionable text commands by the \textbf{text decoder} (right). The shaded area above illustrates the internal structure of the LLAMA3 attention block.
	}
	
	\label{fig_5}
\end{figure*}

\subsection{Supervisory Feedback via SLM}

To address hallucinations and logical inconsistencies in Vision-Language Models (VLMs) for robotic control tasks, this study proposes a supervisory feedback mechanism based on a Small Language Model (SLM). The mechanism ensures the precision, consistency, and safety of control commands through iterative optimization. In practice, we identify three primary types of errors: parameter errors, logical errors, and constraint violations, which involve control values exceeding operational ranges, task execution sequence disorders, and violations of safety and task-specific conditions, respectively. We specifically train the SLM to retain this critical information and validate key parameters in the VLM's output, such as joint angles, end-effector positions, and grip forces, ensuring logical consistency and adherence to environmental constraints. Through continuous queries and verification processes, the SLM leverages its knowledge base to guide the VLM in correcting its outputs, thereby enhancing task success rates.

The overall architecture of the supervisory system is illustrated in Fig.~\ref{fig_5}, revealing the feedback loops and information flow between system components. Subsequent sections will discuss in detail the following four crucial aspects: \textbf{"Model Finetuning"}, \textbf{"SLM Prompt Template Design"}, \textbf{"Supervision Strategy"} and \textbf{"SLM-VLM Interaction Framework"}, all of which are essential for constructing the supervisory system.

\subsubsection{Domain-Specific \textbf{Model Fine-Tuning}}
In practical operational tasks, errors in VLM outputs can be categorized into three primary defined as follows:
\begin{itemize}
	\item \textbf{Parameter Errors}: These occur when control values exceed operational ranges (e.g., joint angles \( \theta \notin [\theta_{\text{min}}, \theta_{\text{max}}] \)).
	\item \textbf{Logical Errors}: These pertain to violations of task sequence constraints (e.g., releasing an object before grasp completion).
	\item \textbf{Constraint Violations}: These refer to breaches of safety or task-specific conditions (e.g., end-effector colliding with a barrier).
\end{itemize}
To optimize the supervisory capabilities of the SLM, we train it specifically to recognize and address these errors. This training involves fine-tuning the SLM using a curated dataset that includes examples of these error types, along with corresponding corrections and confidence scores. 

\paragraph{Dataset}
The fine-tuning process leverages a dataset \(\mathcal{D}\) constructed as follows:
\begin{equation}
	\mathcal{D} = \{(x_i, y_i, c_i)\}_{i=1}^N
\end{equation}
where \(x_i\) represents input commands, \(y_i\) denotes correct outputs, and \(c_i\) indicates confidence scores. It includes four critical components:
\begin{itemize}
	\item Natural-language task descriptions with validated control parameters
	\item Expert-annotated correction examples with confidence scores
	\item Historical execution logs with success/failure cases
	\item Task-specific constraints and safety thresholds
\end{itemize}
The construction details are presented in Section~\ref{sec:experiments}. 

\paragraph{Fine-Tuning Method}
The fine-tuning process leverages \textbf{Low-Rank Adaptation (LoRA)}, which introduces trainable matrices \(A \in \mathbb{R}^{r \times d}\) and \(B \in \mathbb{R}^{d \times r}\) with rank \(r\) into the pre-trained weights \(W\), as shown below:
\begin{equation}
	W + BA = W + \Delta W
\end{equation}
This parameter-efficient adaptation balances domain specificity and computational efficiency~\cite{ref36-lora}. The rank \(r\) controls the adaptation capacity, with lower values preserving more original knowledge. During fine-tuning, a cross-entropy loss function is employed to minimize prediction errors while preserving logical coherence in task outputs. Critical hyperparameters, such as rank and learning rate, are empirically optimized.

\subsubsection{\textbf{SLM Prompt Template} for SLM--VLM Interaction}

To standardize the supervision process, we define the \textbf{SLM Prompt Template} structure, see TABLE \ref{tab:slm_prompt_template}. This template provides necessary information to the SLM for generating accurate feedback and adjustment suggestions. Details are presented in Section \ref{sec:experiments}.

\begin{table}[h!]
	\centering
	\caption{SLM--VLM Interaction Prompt Template}
	\label{tab:slm_prompt_template}
	\begin{tabularx}{0.5\textwidth}{@{\extracolsep{\fill}}p{0.2\textwidth}X}
		\toprule
		\textbf{Component} & \textbf{Description} \\
		\midrule
		\textbf{Input} & \\
		Task Description (\{T\}) & The task description provided to the system. \\
		Prompt Template (\{PT\}) & Template used to structure prompts for VLM input. \\
		Historical Responses (\{H$^{(n)}$\}) & Record of prior interactions and outputs for context. \\
		VLM Output (\{R$^{(n)}$\}) & The current output generated by the VLM. \\
		\midrule
		\textbf{Instruction} & \\
		Validate & Verify parameters, logic, and constraints in the VLM output. \\
		Identify Issues & Identify any inconsistencies or errors in the output. \\
		Provide Suggestions & Offer actionable recommendations with confidence scores. \\
		\midrule
		\textbf{Output} & \\
		Feedback (\{F\}) & Constructive feedback for improving the VLM output. \\
		Adjustments (\(\Delta\)) & Suggested changes to refine the output. \\
		Confidence (\(\sigma\)) & Confidence level associated with feedback and adjustments. \\
		\bottomrule
	\end{tabularx}
\end{table}

\subsubsection{\textbf{Supervision Strategies} for SLM-VLM Interaction}
To ensure stability and convergence within the feedback loop, the SLM employs two complementary strategies:
\paragraph{Single-Dimension Adjustments} 
This strategy focus each iteration on refining one parameter or logical step at a time, thereby avoiding interference among interdependent variables:
\begin{equation}
	\Delta p_j^{(n)} = \delta_{j, i(n)} \cdot f\left(e_j^{(n)}\right)
\end{equation}
where \( i(n) \) denotes the parameter selected for adjustment in the \( n \)-th iteration, \( p_j \) represents the \( j \)-th parameter, and \( e_j \) is its associated error. This multi-stage optimization ensures monotonic improvement through incremental parameter updates.

\paragraph{Feedback History Tracking}
This strategy maintains a record of past adjustments, preventing redundant or contradictory corrections:
\begin{equation}
	\mathcal{H}^{(n)} = \left\{\Delta p_i^{(k)} \,|\, k = 1, \ldots, n-1\right\}
\end{equation}
Here, \( \mathcal{H}^{(n)} \) represents the feedback history. Given that the SLM is a small model, it is solely responsible for adjusting the VLM's outputs in complex tasks. By caching historical adjustments, the SLM can quickly provide optimal feedback for tasks it has previously corrected, thereby enhancing efficiency.

\paragraph{Implementation Example} 
Consider a VLM-generated command for robotic manipulation:
\begin{quote}
	\textit{"Apply 15\,N grip force on fragile component"}
\end{quote}
The SLM identifies a safety violation (max 5\,N threshold) and generates corrective feedback:
\begin{quote}
	\emph{``A grip force of 15\,N exceeds the safety threshold for fragile components (maximum allowable force: 5\,N). Please adjust the grip force to at most 5\,N and reattempt the grasping action while maintaining the current approach vector and speed.''}
\end{quote}
This feedback is reintegrated into the text prompt helping VLM to regenerate commands until either the commands satisfy all task requirements or a predefined iteration limit \( N_{\text{max}} \) is reached. 
If convergence remains unattainable after \( N_{\text{max}} \) iterations, a fallback mechanism ensures task safety by either reverting to a default safe state or requesting human intervention, depending on the task's criticality level.

\begin{algorithm}[t]
	\caption{SLM--VLM Interaction Process}
	\label{alg:slm_vlm_interaction}
	\begin{algorithmic}[1]
		\STATE \textbf{Input:} Task description \(T\), prompt template \(PT\), maximum iterations \(N_{\text{max}}\), confidence threshold \(\tau\), historical responses \(H^{(n)}\), VLM output \(R^{(n)}\) (with confidence score \(C^{(n)}\)), acceptance flag \(flag^{(n)} \in \{0,1\}\), issue set \(I^{(n)}\), suggestion set \(S^{(n)} = \{(s_i, c_i)\}\) where \(c_i \in \{A,B,C\}\), and VLM prompt \(P^{(n)}\)
		\STATE \textbf{Output:} Final robotic command \(R^{*}\), Recorded information \(R\)
		\STATE Initialize \(n \leftarrow 0\), \(H^{(0)} \leftarrow \emptyset\), \(R^{(0)} \leftarrow \emptyset\)
		\WHILE{ \(n < N_{\text{max}} \) }
		\STATE \textbf{SLM Processing:}
		\STATE \quad Generate feedback \(Y^{(n)} = \{\text{flag}^{(n)}, C^{(n)}, I^{(n)}, S^{(n)}, \newline
		\hspace*{6.5em} P^{(n)}\}\) based on \(T\), \(PT\), \(H^{(n)}\), and \(R^{(n)}\)
		\STATE \quad Update history \(H^{(n+1)} = H^{(n)} \cup \{Y^{(n)}\}\)
		\STATE \textbf{VLM Processing:}
		\STATE \quad Generate output \(R^{(n+1)} = \text{VLM}(P^{(n)})\)
		\STATE \textbf{Evaluation:}
		\IF{ \(C^{(n+1)} > \tau\) \textbf{and} \(\text{flag}^{(n+1)} = 1\) }
		\STATE \textbf{Finalize:}
		\STATE \quad Require VLM to describe the scenario in detail.
		\STATE \quad Adopt VLM's output \(R^{(n+1)}\) for robot control.
		\STATE \quad Record \(R = \{\text{Scenario Info}, \text{Task Info}, \newline 
		\hspace*{6.5em} \text{Robot Control Code}\}\)
		\STATE \quad \textbf{Terminate}
		\ENDIF
		\IF{ \(n+1 = N_{\text{max}} \) }
		\STATE \textbf{Fallback Mechanism:}
		\STATE \quad Revert to \(H^{(n)}\)
		\STATE \quad Identify the action with the lowest confidence level
		\STATE \quad Generate a correction prompt for this action
		\STATE \quad \textbf{Continue} to next iteration
		\ENDIF
		\STATE Increment \(n \leftarrow n + 1\)
		\ENDWHILE
		\STATE \textbf{Task Outcome:}
		\IF{ Task \textbf{Successful} }
		\STATE Record \(R = \{\text{Scenario Info}, \text{Task Info}, \newline
		\hspace*{6.5em} \text{Final Robot Control Code}\}\)
		\ELSE
		\STATE Record failure and request human intervention
		\ENDIF
	\end{algorithmic}
\end{algorithm}

\subsubsection{\textbf{SLM-VLM Interaction Framework}}
The interaction between the \textbf{Small Language Model (SLM)} and \textbf{Vision--Language Model (VLM)} constitutes a closed-loop feedback system for iterative refinement of robotic control commands. This dual-model approach enables robust command generation through continuous validation and refinement. See Algorithm\ref{alg:slm_vlm_interaction}.

\paragraph{System Formalization}
The interaction is formalized as a state transition system \(\mathcal{S} = (X, U, F)\), where \(X \subseteq \mathbb{R}^n\) stands for the state space of possible commands, \(U \subseteq \mathbb{R}^m\) is the feedback space and \(F: X \times U \rightarrow X\) refers to state transition function.

\paragraph{Interaction Components}
The SLM generates feedback \(Y^{(n)}\) at iteration \(n\):
\[
Y^{(n)} = \{\text{flag}^{(n)}, C^{(n)}, I^{(n)}, S^{(n)}, P^{(n)}\}
\]
where \(Y^{(n)}\) represents the feedback generated by the SLM in the \(n\)-th iteration, encompassing the acceptance flag, confidence score, detected issues, corresponding suggestions, and the prompt for the VLM.

\begin{figure*}[ht]
	\centering
	\includegraphics[width=\textwidth, trim={40px} {40px} {40px} {90px}, clip]{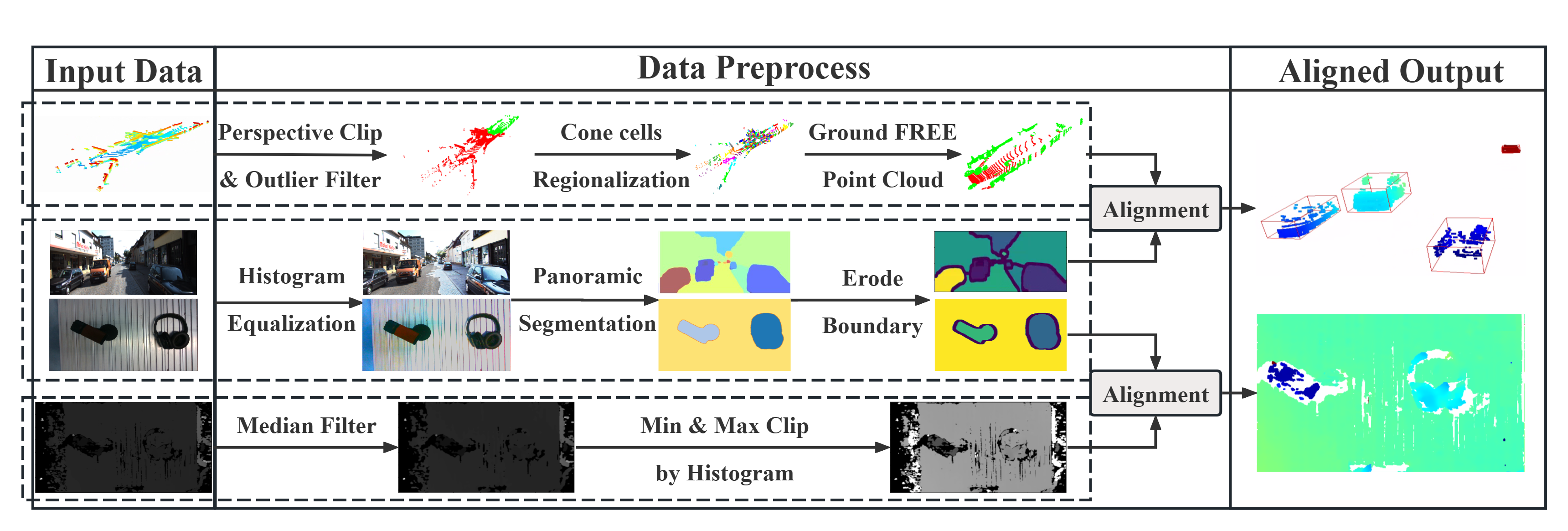}
	\caption{
		The complete data preprocessing pipeline and experiment results. Inputs (left) include point cloud data, RGB images, and depth maps. The central process (middle) consists of three layers: \textbf{Point Cloud Processing} (top), \textbf{Image Processing} (center), and \textbf{Depth Map Processing} (bottom). Image processing further distinguishes KITTI dataset images (top row) for point cloud alignment and real-world robotic arm images (bottom row) for depth map alignment. Outputs (right) are the \textbf{aligned and segmented point cloud} and \textbf{aligned depth map}.
	}
	
	\label{fig_6}
\end{figure*}

\section{Experiments}
\label{sec:experiments}
In this section, we evaluate the effectiveness of the proposed modular framework in enhancing scene understanding and task execution for robotic systems. We focus on assessing the contributions of the fusion and supervision modules in terms of 3D spatial reasoning, task planning, and control code generation.

\subsection{Experimental Setup}
We conducted experiments using a FRANKA robotic arm equipped with a RealSense RGB-D camera and a LiDAR sensor. The robot performed a series of manipulation tasks in an indoor environment featuring moderate clutter and varying lighting conditions. Camera and LiDAR data were used for scene perception and task planning.

\subsubsection{Camera and LiDAR Registration}
Accurate alignment of camera and LiDAR data was achieved using pre-calibrated intrinsic and extrinsic parameters. A transformation matrix \(T_{\text{Lidar}}^{\text{Camera}}\) was computed to project LiDAR points into the camera frame, enabling precise multimodal data fusion for 2D prompt synthesis. The alignment accuracy was verified by comparing projected point cloud edges with image contours, ensuring consistent integration of RGB and depth information.
\begin{equation}
	\begin{bmatrix}
		X_{\text{camera}} \\
		Y_{\text{camera}} \\
		Z_{\text{camera}} \\
		1
	\end{bmatrix}
	=
	T^{\text{camera}}_{\text{Lidar}} \cdot
	\begin{bmatrix}
		X_{\text{Lidar}} \\
		Y_{\text{Lidar}} \\
		Z_{\text{Lidar}} \\
		1
	\end{bmatrix}
	\label{eq:transform}
\end{equation}

\begin{equation}
	\begin{bmatrix}
		u \\
		v \\
		1
	\end{bmatrix}
	=
	\frac{1}{Z_{\text{camera}}}
	\begin{bmatrix}
		f_x & 0 & c_x \\
		0 & f_y & c_y \\
		0 & 0 & 1
	\end{bmatrix}
	\begin{bmatrix}
		X_{\text{camera}} \\
		Y_{\text{camera}} \\
		Z_{\text{camera}}
	\end{bmatrix}
	\label{eq:projection}
\end{equation}

\subsubsection{Data Preprocessing}
High-quality data preprocessing is crucial for robust multimodal fusion. The following preprocessing steps were applied to point cloud, image, and depth map data:

\begin{itemize}
	\item \textbf{Point Cloud Processing}: The cone cell-based segmentation method divides the point cloud within the camera's frustum into multiple regions, effectively addressing issues caused by multiple ground layers\cite{ref44-patchwork++}. PCA and thresholding are applied within each region to remove ground points. Additionally, redundancy sampling is employed to reduce computational overhead.
	
	\item \textbf{Image Processing}: Histogram-based color equalization, viewpoint cropping, and neural network-based segmentation techniques are used for panoramic segmentation, improving the clarity and quality of the images.
	
	\item \textbf{Depth Map Processing}: Filtering and cropping methods are applied to reduce artifacts and ensure smooth depth transitions.
\end{itemize}

These steps, as illustrated in Figure \ref{fig_6}, ensure that the fused data is accurate and robust for 2D prompt synthesis, with corresponding experimental results shown in the same figure.

\subsection{Prompt Template}
In this experiment, prompt templates play a critical role in guiding the Vision-Language Model (VLM) and Small Language Model (SLM) to perform robotic task planning and validation. These templates standardize input-output interactions, ensuring task-specific requirements are met while addressing logical consistency and safety constraints. The VLM template focuses on generating task steps from visual and textual inputs, while the SLM template supervises the VLM outputs and provides necessary adjustments. The following sections detail the configuration of these templates.
\subsubsection{VLM Template}
The Vision-Language Model (VLM) is tasked with generating task plans based on visual and textual inputs. Its output directly influences the robotic actions, making it critical to standardize the input and output structure for consistent performance. This template provides the VLM with task descriptions, prior knowledge of the robot and the environment, and necessary constraints, ensuring safe and precise task execution. See \ref{VLM-Prompt}.

\subsubsection{SLM Template}
The Small Language Model (SLM) ensures that the VLM outputs adhere to logical consistency, safety, and task-specific requirements. Acting as a validation layer, the SLM identifies potential issues in the VLM outputs and provides corrective feedback to refine the task plan. The template defines the input-output structure and specifies the feedback format to guide the VLM iteratively. See \ref{SLM-Prompt}.

\subsection{Output Examples}
This section provides example outputs from the Vision-Language Model (VLM) and the Small Language Model (SLM) for a specific robotic task. These outputs illustrate how the models interact and refine their responses during task execution. The task is defined as follows:
\begin{quote}
	\textit{Task Description: Identify and remove a headphone from a headphone stand in the scene, and place it into the user's hand. Prompt Template: \{PT\}, Historical Responses: \{H$^{(n)}$\}, VLM Output: Grasp the headphone.}
\end{quote}
The following subsections detail the outputs from the VLM and SLM for this task, demonstrating the iterative refinement process.

\subsubsection{VLM Output}
An example output is stated in \ref{VLM-Output} generated by the Vision-Language Model (VLM) after processing the task description and environmental inputs. The VLM attempts to generate a sequence of robotic actions to achieve the task. Output highlights the VLM's initial attempt to generate a valid sequence of robotic actions. However, certain issues, such as the use of undefined functions and inappropriate parameters, require further refinement.

\subsubsection{SLM Output}
The Small Language Model (SLM) analyzes the VLM's output and identifies areas for improvement. It provides feedback and suggestions to enhance the logical consistency and safety of the action sequence. \ref{SLM-Output} is an example of the SLM's feedback.
The example demonstrates how the SLM enhances the action plan by addressing key issues and ensuring that the final commands are safe and executable. The iterative feedback loop between the VLM and SLM ensures the task's successful completion.

\subsection{SLM Dataset Construction}
\label{data_format}

To train the Small Language Model (SLM), we constructed a dataset combining public datasets, custom experimental data, and augmented data. All raw data were processed to conform to the SLM's input-output format, ensuring compatibility with next token prediction training.

\subsubsection{Positive Data Collection}  
We utilized \textbf{BridgeData V2}\cite{ref37-dataset}, focusing on task descriptions and task decompositions while removing trajectory-related data. From this dataset, we extracted \textbf{240 samples}, representing \textbf{24 unique scenes}, with 10 frames uniformly sampled from each scene. Each sample includes a task description, task decomposition, and object coordinates within the scene. Additionally, we collected \textbf{320 task samples} in a controlled environment featuring a robotic arm, headphone stand, and headset. Object positions were systematically varied to generate scene descriptions with precise 3D coordinates relative to the robot base, along with corresponding task decompositions.

\subsubsection{Data Augmentation and Negative Data Generation}  
To diversify the dataset and simulate challenging scenarios, we generated \textbf{3,000 augmented samples}. The augmentation strategies included:
\begin{itemize}
	\item \textbf{Positive Data Augmentation:}
	\begin{itemize}
		\item Modifying task parameters to exceed operational thresholds (e.g., increasing grip force beyond safe limits).
		\item Removing or adding steps in task decomposition to simulate incomplete or invalid sequences (e.g., omitting the release step in a pick-and-place task).
	\end{itemize}
	\item \textbf{Negative Data Generation:}
	\begin{itemize}
		\item Reversing task flows, such as swapping grasp and release actions.
		\item Introducing parameter violations, such as invalid object positions outside the robot’s workspace.
		\item Simulating occlusions by removing parts of the object’s location data in the scene description.
	\end{itemize}
\end{itemize}

Each augmented sample was paired with its predefined error type, and GPT-4 was tasked with generating feedback strictly based on the given error and task information. This ensured that the feedback aligned with the predefined errors while conforming to the SLM input-output format.

\subsubsection{Data Processing and Final Dataset Composition}
All collected and generated data were initially raw data and were transformed into complete input-output pairs to prepare the dataset for next token prediction training. Each sample follows the input-output structure detailed in Appendix \ref{SLM-Prompt}. The final dataset comprises \textbf{240 samples} from BridgeData V2, \textbf{320 samples} from custom experiments, and \textbf{3,000 augmented samples}. All samples are stored in a structured JSON format compatible with the SLM training pipeline, balancing positive and negative samples to ensure effective learning of both task validation and error correction.

\subsection{Evaluation Tasks}
\label{task}
The proposed framework was evaluated on four manipulation tasks using the FRANKA robotic arm equipped with an RGB-D camera. These tasks were designed with increasing complexity to assess the framework's performance in scene understanding, task planning, and control code generation:

\begin{itemize}
	\item \textbf{Task 1}: Hanging a headphone on a stand. The task begins with the headphone placed on a flat surface and the stand fixed at a predefined location. This task evaluates the system's spatial perception and object manipulation capabilities.
	
	\item \textbf{Task 2}: Placing the headphone in a designated position. Starting with the headphone hanging on the stand, the robot must identify and remove it, then place it at a specified 3D location. This task tests the framework's ability to perform object segmentation, spatial coupling, and precise placement.
	
	\item \textbf{Task 3}: Moving the stand and hanging the headphone. The robot must first relocate the stand to a specified position and then hang the headphone on it. This task evaluates the framework’s dual-task reasoning, coordination, and sequential execution capabilities.
	
	\item \textbf{Task 4}: Generating control code from high-level instructions (e.g., "listen to music"). The robot must interpret the instruction, generate executable control commands, and perform the corresponding actions, such as placing the headphone in the appropriate position. This task evaluates the framework's ability to translate abstract commands into logically consistent and executable actions.
\end{itemize}

\subsection{Evaluation Metrics}
To evaluate the proposed framework, four key metrics were used: \textbf{mIoU}, \textbf{ROUGE-L}, \textbf{Executability}, and \textbf{TSR}. These metrics assess spatial localization accuracy, task decomposition ability, command executability, and overall task success rate.

\subsubsection{Mean Intersection over Union (mIoU)} 
mIoU measures the localization accuracy by comparing the overlap between the predicted and ground-truth object regions in the 3D space. The formula is:
\begin{equation}
	\text{mIoU} = \frac{1}{N} \sum_{i=1}^{N} \frac{|A_i \cap B_i|}{|A_i \cup B_i|},
\end{equation}
where $A_i$ and $B_i$ represent the predicted and ground-truth points, respectively, constrained by a 3D localization error of less than 0.2. This adaptation ensures mIoU evaluates the spatial precision required for accurate object manipulation.

\subsubsection{ROUGE-L}
ROUGE-L evaluates the semantic consistency of generated control sequences by computing the longest common subsequence (LCS) between the predicted sequence $P$ and the ground-truth sequence $G$:
\begin{equation}
	\text{ROUGE-L} = \frac{2 \cdot LCS(P, G)}{|P| + |G|}.
\end{equation}
This metric focuses on high-level task command consistency, such as evaluating instructions like "listen to music."

\subsubsection{Executability}
Executability quantifies the proportion of control commands that can be successfully parsed and executed by the robot:
\begin{equation}
	\text{Executability} = \frac{\text{Successful Executions}}{\text{Total Executions}} \times 100\%.
\end{equation}
Parsing ensures that each generated command conforms to the robot's syntax and functional constraints, reflecting the compatibility of instructions with physical capabilities.

\subsubsection{Task Success Rate (TSR)}
TSR measures the overall success rate of task completion:
\begin{equation}
	\text{TSR} = \frac{\text{Successful Tasks}}{\text{Total Tasks}} \times 100\%.
\end{equation}
A task is considered successful only if all steps in the generated plan are executed without failure. Repeated trials ensure that TSR captures the robustness and reliability of the framework under real-world conditions.

\begin{figure*}[ht]
	\centering
	\includegraphics[width=\textwidth, trim={40px} {40px} {40px} {40px}, clip]{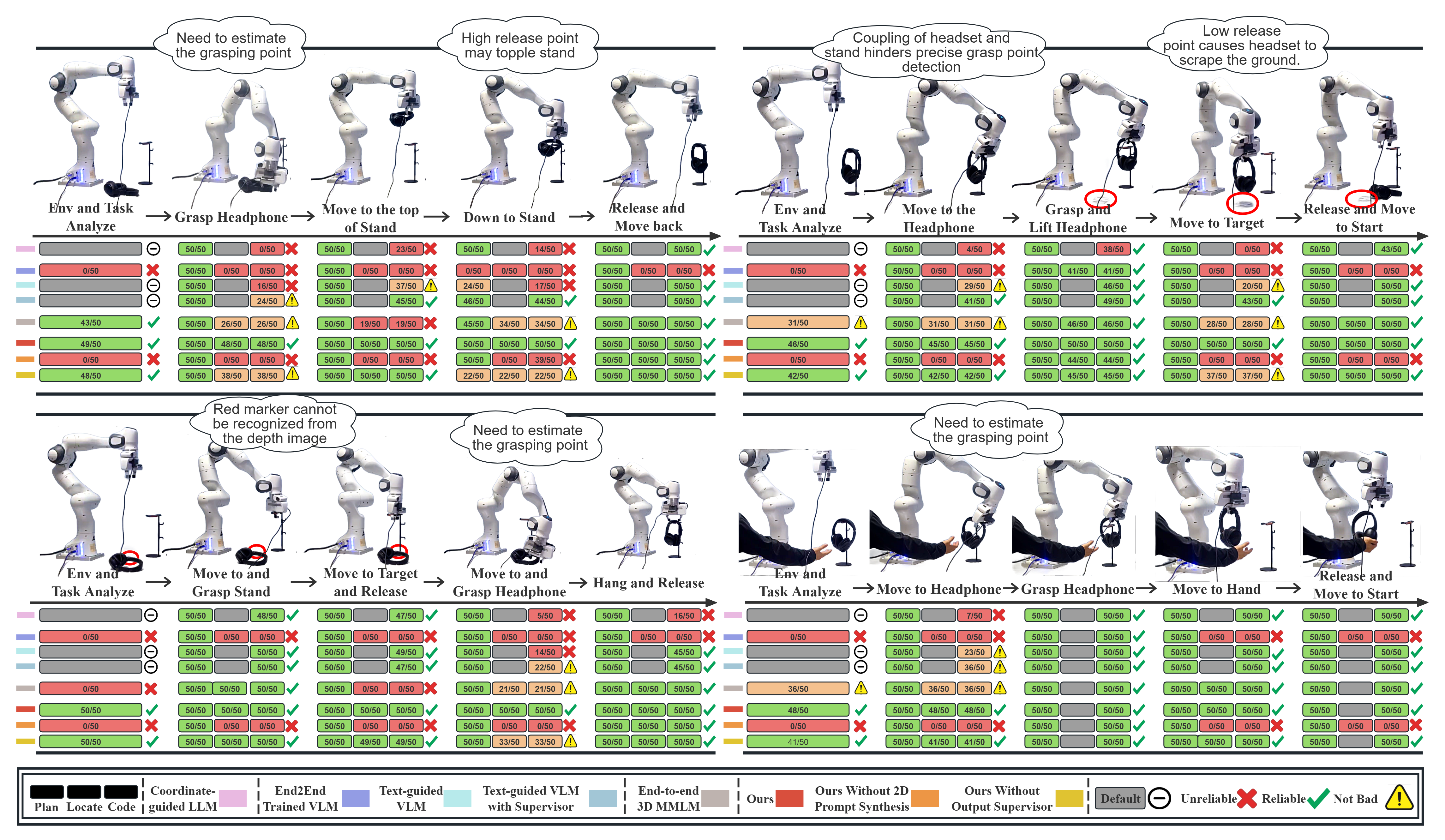}
	\caption{
		 Multi-Dimensional Performance Evaluation of Robotic Headphone Manipulation Tasks. The figure is divided into four quadrants, each representing a distinct headphone manipulation task mentioned in \ref{task}. Each quadrant comprises two sections. The upper section displays five sequential real-time operational states of the robotic arm, each accompanied by corresponding action descriptions (e.g., "Grasp headphone"). The lower section contains five 3×8 tables, each corresponding to one of the five operational states. Within each table, the three columns represent distinct evaluation criteria: (1) success of task step planning, (2) accuracy of position recognition, and (3) executability of the generated control code. Each table evaluates eight different robotic learning models, including Coordinate-guided LLM, End-to-End Trained VLM, Text-guided VLM variants, 3D Multimodal Learning Models, and the proposed approaches with and without specific prompt synthesization and supervisory mechanisms. Performance metrics are visualized through color-coded cells based on success rates from 50 experimental trials: red (0–40\% reliability, unreliable), yellow (40–80\% reliability, not bad), and green (80–100\% reliability, reliable). A legend at the bottom of the figure elucidates the model names and the significance of the color codes.
	}
	
	\label{fig_prompt_comparison}
\end{figure*}

\begin{table*}[ht]
	\caption{Performance Comparison of Various Models Across Four Tasks}
	\label{table:performance_comparison}
	\centering
	\resizebox{\textwidth}{!}{%
		\begin{tabular}{m{4cm}cccccccccccccccc}
			\toprule
			\multirow{2}{*}{\makecell{\textbf{Model Name}}} & \multicolumn{4}{c}{\textbf{Task 1}}               & \multicolumn{4}{c}{\textbf{Task 2}}               & \multicolumn{4}{c}{\textbf{Task 3}}               & \multicolumn{4}{c}{\textbf{Task 4}}               \\ \cmidrule(lr){2-5} \cmidrule(lr){6-9} \cmidrule(lr){10-13} \cmidrule(lr){14-17}
			& \textbf{mIOU↑} & \textbf{R-L↑} & \textbf{Exe↑} & \textbf{TSR↑} & \textbf{mIOU↑} & \textbf{R-L↑} & \textbf{Exe↑} & \textbf{TSR↑} & \textbf{mIOU↑} & \textbf{R-L↑} & \textbf{Exe↑} & \textbf{TSR↑} & \textbf{mIOU↑} & \textbf{R-L↑} & \textbf{Exe↑} & \textbf{TSR↑} \\ \midrule
			Text-guided LLM                                  & --            & 0.683        & 0.435        & 0            & --            & 0.590        & 0.425        & 0            & --            & 0.365        & 0.580        & 0.029        & --            & 0.590        & 0.785        & 0.140        \\
			End2End VLM                                      & 0             & 0.780        & 0            & 0            & 0             & 0.680        & 0.205        & 0            & 0             & 0.432        & 0            & 0            & 0             & 0.680        & 0.250        & 0            \\
			Text-guided VLM                                  & --            & 0.780        & 0.600        & 0.081        & --            & 0.680        & 0.725        & 0.213        & --            & 0.432        & 0.790        & 0.247        & --            & 0.680        & 0.865        & 0.460        \\
			Text-guided VLM (COT)                            & --            & 1.00         & 0.815        & 0.380        & --            & 0.880        & 0.915        & 0.691        & --            & 0.625        & 0.820        & 0.372        & --            & 0.880        & 0.930        & 0.720        \\
			3D Multimodal LM                                 & 0.525         & 0.780        & 0.645        & 0.134        & 0.627         & 0.680        & 0.775        & 0.319        & 0.567         & 0.432        & 0.605        & 0            & 0.781         & 0.680        & 0.930        & 0.740        \\
			\textbf{Ours}                                    & \textbf{0.981} & \textbf{1.00} & \textbf{0.990} & \textbf{0.960} & \textbf{0.955} & \textbf{1.00} & \textbf{0.975} & \textbf{0.900} & \textbf{0.990} & \textbf{1.00} & \textbf{1.00}  & \textbf{1.00} & \textbf{0.962} & \textbf{1.00} & \textbf{0.990} & \textbf{0.960} \\
			Ours (-2D Prompt)                                & 0             & 1.00         & 0.445        & 0            & 0             & 0.880        & 0.220        & 0            & 0             & 0.625        & 0.250        & 0            & 0             & 0.880        & 0.250        & 0            \\
			Ours (-SLM)                                      & 0.723         & 0.780        & 0.800        & 0.334        & 0.782         & 0.680        & 0.870        & 0.559        & 0.863         & 0.432        & 0.910        & 0.647        & 0.847         & 0.680        & 0.955        & 0.820        \\ \bottomrule
		\end{tabular}%
	}
\end{table*}

\subsection{Results and Analysis}

Experimental results, summarized in Table~\ref{table:performance_comparison}, highlight the key limitations of existing models and demonstrate the proposed framework's advantages.

\subsubsection{Task Execution Comparison}
The experimental results reveal fundamental limitations across different categories of models, highlighting the advantages of the proposed framework. For models incapable of autonomously acquiring 3D coordinates (e.g., LLM, VLM, VLM with COT), precise object positions were directly provided to simulate scenarios where such information might be obtained via alternative means (e.g., prompt engineering, manual input, or pre-training on specific scenes). This setup effectively isolates the models’ spatial data acquisition capabilities, allowing for a fair and focused evaluation of their spatial understanding, task planning, and execution performance. It should be noted, however, that providing spatial data does not equate to enhancing the models' inherent spatial reasoning capabilities, although the use of prompts indirectly facilitates task understanding to some extent.

\begin{itemize}
	\item \textbf{Coordinate-Guided Models (e.g., LLAMA3.3 70B)}:  
	These models, rooted in high-dimensional text embedding spaces, lack the capacity to autonomously integrate spatial data into coherent 3D environment representations. Even with precise object coordinates provided, their inability to understand spatial relationships results in execution errors. For example, in Task 1 (hanging the headphone) and Task 3 (moving and hanging the headphone), these models fail to predict spatial constraints, frequently causing collisions. This limitation reflects the inherent disconnect between text-based reasoning and physical task requirements in dynamic, multi-object scenarios.
	
	\item \textbf{End-to-End Vision-Language Models (e.g., GPT-4o)}:  
	While incorporating 2D visual inputs improves task planning, these models exhibit unclear understanding of spatial coupling relationships. In Task 2 (placing the headphone), for instance, the lack of clear comprehension of spatial coupling leads to inconsistent object placement and suboptimal task execution. This highlights a key limitation in their ability to extrapolate 3D spatial relationships from 2D inputs, particularly in unstructured environments.
	
	\item \textbf{Text-Guided Reasoning Models (e.g., GPT-o1 with COT)}:  
	Incorporating chain-of-thought reasoning improves task decomposition and planning, demonstrating better performance compared to standard vision-language models. This feedback-based and iterative reasoning approach enables deeper consideration of task constraints, particularly in tasks requiring spatial depth reasoning. However, despite the provision of precise coordinates, these models lack adaptive refinement during task execution. For instance, in Task 3, their inability to dynamically adjust plans results in frequent execution errors in object interactions, reflecting an over-reliance on static reasoning.
	
	\item \textbf{3D Multimodal Learning Models (e.g., 3D-LLM)}:  
	These models integrate multimodal inputs, achieving higher spatial reasoning capabilities compared to text-only or vision-language models. However, their reliance on static, highly precise inputs reduces their robustness in real-world environments. For instance, in Task 4 (generating control code), these models struggle to adapt to environmental ambiguities or dynamically changing contexts, often producing incomplete or erroneous task plans. This highlights their difficulty in generalizing to unstructured or dynamic operational conditions.
	
	\item \textbf{Proposed Framework}:  
	The proposed framework demonstrates significant advantages by autonomously acquiring and processing spatial data through the \textbf{2D Prompt Synthesis Module} and ensuring task consistency via the \textbf{Output Supervision Module}. These components enable the framework to construct rich, context-aware spatial representations, iteratively refine task plans, and adapt to dynamic scenarios. For example, in Task 3, the framework dynamically adjusted object positions, achieving \textbf{100\% TSR} with precise localization and collision-free execution. Furthermore, in Task 4, the framework translated high-level commands into detailed task plans with \textbf{ROUGE-L = 1.00}, demonstrating logical consistency and high executability. These results highlight the framework's robustness in handling spatially complex and semantically rich tasks.
\end{itemize}

This analysis focuses on the key limitations of existing model categories while providing a clear evaluation of the proposed framework’s performance.

\subsubsection{Ablation Study}
The ablation study, detailed in the last three rows of Table~\ref{table:performance_comparison}, evaluates the individual contributions of the \textbf{2D Prompt Synthesis Module} and \textbf{Output Supervision Module}.  Results validate their necessity for robust spatial reasoning and task execution.

\paragraph{2D Prompt Synthesis Module}  
Removing this module resulted in a complete failure across all tasks (\textbf{0\% TSR}). This highlights its indispensable role in embedding spatial details into task planning. Without this module:
\begin{enumerate}
	\item \textbf{Object Relationship Errors:} In spatially complex tasks, such as Task 1 (hanging the headphone) and Task 3 (moving and hanging the headphone), the framework misinterpreted object relationships, leading to failed grasping or frequent collisions.
	\item \textbf{Trajectory Generation Failure:} The absence of spatial prompts rendered the framework incapable of generating accurate object trajectories, particularly in dynamic scenarios. 
\end{enumerate} 
This demonstrates that the 2D Prompt Synthesis Module provides essential geometric priors for environments with coupled objects or moving targets.
	
\paragraph{Output Supervision Module}  
Eliminating this module caused substantial performance degradation, with TSR dropping by \textbf{67\%} in Task 2 and ROUGE-L decreasing by \textbf{22\%}. This module serves as a critical safeguard for ensuring logical consistency and reducing hallucinations in generated control sequences. Without it:
\begin{enumerate}
	\item \textbf{Task Planning Hallucinations:} In Task 2 (placing the headphone), the framework often produced infeasible or incorrect task plans, such as placing the headphone outside the target zone.
	\item \textbf{Sequence Incoherence:} In Task 4 (high-level command execution), hallucinations in task decomposition increased, leading to incomplete or logically inconsistent sequences. 
\end{enumerate}
These results highlight the Output Supervision Module’s importance in validating control commands against spatial and logical constraints, ensuring reliable execution across diverse tasks.

\paragraph{Conclusion} The ablation study reveals how these two modules jointly address limitations inherent to existing models. The \textbf{2D Prompt Synthesis Module} bridges the gap in spatial reasoning by embedding precise contextual information into task planning, while the \textbf{Output Supervision Module} ensures semantic and logical coherence in task execution. Together, they enable the proposed framework to perform reliably in spatially challenging and semantically complex tasks, achieving consistent success rates even in dynamic and ambiguous environments.

\subsubsection{Discussion and Insights}

This section delves deeper into the experimental results, analyzes the strengths and limitations of the proposed framework, and discusses future research directions. The experiments aim to validate the effectiveness of the framework in autonomous 3D spatial understanding, efficient task planning based on VLM and Chain-of-Thought (COT), and improving control sequence reliability through the Output Supervision Module. The results (Table~\ref{table:performance_comparison}) clearly demonstrate the superiority of our framework in various performance metrics, particularly Task Success Rate (TSR) and spatial localization accuracy (mIoU).

\paragraph{Autonomous 3D Data Acquisition and Spatial Understanding}

A key innovation of our framework is the ability to autonomously acquire and process 3D spatial information, in contrast to traditional methods that rely on manually or pre-defined 3D coordinates. The 2D Prompt Synthesis Module enables effective mapping from 2D image data to 3D space by aligning the image pixels with the point cloud data from a LiDAR sensor. This process allows each pixel to be assigned a 3D coordinate, making real-time, dynamic spatial understanding possible, which is essential for robust task execution in dynamic environments.

However, while our framework performs well in static environments, we observe that challenges arise in dynamic environments, especially due to sensor noise, rapid motion, occlusion, and the inherent limitations of geometry-based registration methods like ICP. These factors can negatively impact the precision of point cloud alignment, thus affecting the accuracy of the 3D coordinates. Future research could focus on exploring more advanced sensor fusion techniques and improving the robustness of point cloud processing to better handle these dynamic challenges.

\paragraph{Task Planning and Execution through Chain-of-Thought (COT)}

The integration of Chain-of-Thought (COT) reasoning into the Text-Guided VLM significantly enhances task decomposition and the logical consistency of generated task plans. For instance, in Task 4, the model demonstrated improved abilities in generating task plans and control sequences that follow logical structures, leading to higher task execution success rates. COT reasoning enables the model to make sequential decisions, which helps to handle task dependencies more effectively. This finding underscores the value of iterative reasoning in improving task execution in complex, dynamic environments.

Despite these improvements, there are still limitations in terms of the model's ability to handle long sequences of complex tasks, especially when dependencies between actions are highly intricate. Further exploration is needed to optimize the reasoning process, such as through multi-step thinking or enhancing the model’s ability to predict future states based on past actions.

\paragraph{Efficiency of the Proposed Framework with Minimal Training Data}

Another significant advantage of our framework is its ability to achieve excellent performance with a minimal amount of training data, unlike other approaches that require large, task-specific datasets. Many existing models are trained on extensive datasets specific to particular tasks or environments, which results in a lack of generalization and inefficiency in resource use. In contrast, our modular approach, inspired by human cognitive processing, leverages the pre-trained capabilities of VLMs, such as image analysis and logical reasoning, to process diverse tasks without retraining on large, task-specific datasets.

This approach not only makes our model more resource-efficient but also improves its generalization capabilities across a wide range of tasks. By incorporating a simple 3D coordinate synthesis head and an output supervision module, we can adapt the model to different environments and tasks with minimal additional training. This highlights the potential of modular, flexible systems in real-world robotic applications, where task-specific data collection and retraining are costly and time-consuming.

\paragraph{Challenges in Dynamic Environment Adaptation}

While the framework excels in static environments, real-time adaptation to dynamic environments remains a significant challenge. The current approach to 3D data acquisition, though effective, relies on point cloud and image alignment, which can be influenced by various environmental factors such as motion blur, occlusion, or sensor misalignment. The ability to adjust to changing environments in real-time, with continuous updates to task plans and spatial data, is a crucial area for future improvement.

This challenge can be further explored through the integration of advanced sensor fusion techniques that combine the strengths of multiple sensors (e.g., RGB-D cameras, LiDAR, and IMUs) to achieve more robust environmental understanding. Moreover, continuous learning techniques, where the model updates its spatial understanding based on new sensory input, could improve real-time adaptability.

\paragraph{Future Research Directions}

Our future work will focus on improving the robustness and accuracy of 3D spatial data acquisition in dynamic environments, particularly by enhancing the alignment techniques used between point clouds and images. Research into more advanced sensor fusion methods and real-time adaptive control strategies will be essential for achieving seamless task execution in ever-changing environments. Additionally, further optimization of the Chain-of-Thought reasoning process will be explored to handle complex, long-term task dependencies more efficiently.

Moreover, we aim to explore the potential of integrating reinforcement learning (RL) to enhance the model’s ability to make adaptive decisions based on trial-and-error, which would be especially useful in environments with high uncertainty. By combining our current framework with RL, we hope to create a more flexible and intelligent robotic system capable of handling a wide range of tasks with minimal supervision.

\section{Conclusion}
This paper introduces a novel framework that enhances Vision-Language Models (VLMs) for precise 3D robotic task execution by integrating multimodal data fusion, 2D prompt synthesis, and Small Language Model (SLM) supervision. Experimental results validate its effectiveness, achieving a \textbf{96.0\% task success rate} in fine-grained operations, such as coupled object handling, and demonstrating robust adaptability across dynamic tasks.While the framework excels in spatial reasoning and logical task planning, its iterative question-answering strategy incurs a computational overhead of \textbf{0.8 seconds per session}, and the current image-depth registration relies on multi-step refinement. Future work will focus on optimizing RGB-depth fusion for real-time efficiency and enhancing robustness under noisy conditions.By addressing these challenges, the framework provides a scalable and cost-effective solution for precision robotic operations, laying a solid foundation for advancements in industrial assembly, autonomous manipulation, and collaborative robotics.

\appendix
\section*{PROMPT \& OUTPUT}
\subsection{VLM Prompt}
\label{VLM-Prompt}
\subsubsection{\textbf{Strategy 1: Direct Task Planning with Predefined Semantic Cues}}

\begin{lstlisting}[style=jsonstyle, caption={Direct Scene Prompt Strategy}]
	Instruction:
	1. You are tasked to generate robotic control steps to complete the given task.
	2. The provided image contains red markers on objects, labeled with their 3D positions relative to the robot base coordinate system.
	3. Use the task description, markers, and any provided SLM Feedback to plan precise and safe actions.
	4. Ensure your output adheres to the following constraints:
	- The robot can perform actions such as move_to, grasp, release, rotate.
	- Avoid collisions with obstacles in the scene.
	- Use safe gripping forces for fragile objects.
	- Operate within the safe zone defined by coordinates: x, y, z in [0, 1].
	Input:
	- Task Description: {T}
	- Image Markers: Red markers indicate object names and positions (e.g., headphone: [x, y, z]).
	- Robot Constraints: 
	- Maximum gripping force: 10 N.
	- Minimum distance from obstacles: 0.1 m.
	- Safe operation zone: x, y, z in [0, 1].
	- SLM Feedback (if any): Feedback from SLM regarding task constraints, logical corrections, or improvements. Example:
	{
		"Feedback": [
		"Reduce gripping force to 5N for fragile objects.",
		"Ensure movement paths avoid obstacle at [0.6, 0.4, 0.3]."
		]
	}
	Output Requirements:
	1. Include "scene_description" to describe object names and positions.
	2. Define "task_steps" with clear and executable actions.
	Output Format:
	{
		"scene_description": {
			"objects": [
			{"name": "object_name", "position": [x, y, z]},
			{"name": "object_name_2", "position": [x2, y2, z2]}
			]
		},
		"task_steps": [
		{"step_id": "1", "action": "move_to([x, y, z])"},
		{"step_id": "2", "action": "grasp('object_name')"}
		]
	}
	
	Example Output:
	{
		"scene_description": {
			"objects": [
			{"name": "headphone", "position": [0.5, 0.3, 0.2]},
			{"name": "stand", "position": [0.4, 0.2, 0.1]}
			]
		},
		"task_steps": [
		{"step_id": "1", "action": "move_to([0.5, 0.3, 0.2])"},
		{"step_id": "2", "action": "grasp('headphone')"},
		{"step_id": "3", "action": "move_to([0.7, 0.3, 0.2])"}
		]
	}
	
\end{lstlisting}

\subsubsection{\textbf{Strategy 2: Iterative Multistep Prompting Strategy}}

\begin{lstlisting}[style=jsonstyle, caption={Iterative Interaction Strategy}]
	Instruction:
	1. You will iteratively generate robotic control steps based on the task and image provided.
	2. If any information is missing or unclear, include these as "issues" and suggest resolutions.
	3. Use historical outputs from previous iterations and any provided SLM Feedback to refine your output.
	4. Ensure your output adheres to the following constraints:
	- The robot can perform actions such as move_to, grasp, release, rotate.
	- Avoid collisions with obstacles in the scene.
	- Use safe gripping forces for fragile objects.
	- Operate within the safe zone defined by coordinates: x, y, z in [0, 1].
	Input:
	- Task Description: {T}
	- Image <I>: Red markers indicate object names and positions.
	- Historical Outputs: 
	- <Iteration 1>: {"task_steps": [...], "issues": [...], "flag": "incomplete"}.
	- <Iteration 2>: {"task_steps": [...], "issues": [], "flag": "complete"}.
	- SLM Feedback (if any): Feedback from SLM regarding task constraints, logical corrections, or improvements. Example:
	{
		"feedback": [
		"Adjust the approach angle to 45 degrees for optimal grasping.",
		"Ensure safe clearance from obstacle at [0.4, 0.3, 0.2]."
		]
	}
	Output Requirements:
	1. Include "scene_description" to describe object names and their positions.
	2. Define "task_steps" with clear and executable actions.
	3. Use "issues" to flag missing or unclear information.
	4. Adjust the "flag" based on the completeness of the output.
	
	Output Format:
	{
		"scene_description": {
			"objects": [
			{"name": "object_name", "position": [x, y, z]}
			]
		},
		"task_steps": [
		{"step_id": "1", "action": "move_to([x, y, z])"},
		{"step_id": "2", "action": "grasp('object_name')"}
		],
		"issues": [
		{"description": "..."}	
		],
		"flag": "default"
	}
	Example Output:
	{
		"scene_description": {
			"objects": [
			{"name": "headphone", "position": [0.5, 0.3, 0.2]},
			{"name": "stand", "position": [0.4, 0.2, 0.1]}
			]
		},
		"task_steps": [
		{"step_id": "1", "action": "move_to([0.5, 0.3, 0.2])"},
		{"step_id": "2", "action": "grasp('headphone')"},
		{"step_id": "3", "action": "move_to([0.7, 0.3, 0.2])"},
		{"step_id": "4", "action": "release()"}
		],
		"issues": [
		{"description": "Gripping force not defined for headphone grasping."}
		],
		"flag": "incomplete"
	}
\end{lstlisting}

\subsection{SLM Prompt}
\label{SLM-Prompt}

\begin{lstlisting}[style=jsonstyle, caption={SLM Prompt Template}, label={lst:filled_slm_prompt_template}]
	Instruction:
	1. You are responsible for evaluating the output generated by the VLM and ensuring its logical consistency, safety, and adherence to the provided task constraints.
	2. Validate the following fields from the VLM output:
	- "scene_description": Ensure all objects and their 3D positions are accurately listed.
	- "task_steps": Check that each step is executable, logically consistent, and safe.
	3. Identify any issues in the output and provide suggestions to resolve them.
	4. If "Historical SLM Feedback" is provided, use it to refine your analysis.
	5. Output your feedback in JSON format, including "Feedback", "Suggestions", "Confidence", and "PromptForVLM".
	Input:
	{
		"Task Description": "Identify and remove a headphone from a headphone stand in the scene, and place it into the user's hand.",
		"VLM Output": {
			"scene_description": {
				"objects": [
				{"name": "headphone", "position": [0.5, 0.3, 0.2]},
				{"name": "stand", "position": [0.4, 0.2, 0.1]}
				]
			},
			"task_steps": [
			{"step_id": "1", "action": "move_to([0.5, 0.3, 0.2])"},
			{"step_id": "2", "action": "grasp('headphone')"},
			{"step_id": "3", "action": "move_to([0.7, 0.3, 0.2])"}
			]
		},
		"Historical SLM Feedback": [
		"In previous tasks, gripping force recommendations for fragile objects were adjusted to 5 N to prevent damage.",
		"Ensure minimum distance from obstacle at [0.6, 0.4, 0.3] in future steps."
		]
	}
	"Output Structure": {
		"Feedback": {
			"Description": "A summary of the main issues identified in the VLM output and their significance.",
			"Example": "The VLM output 'grasp the headphone' lacks specificity regarding the exact gripping position, which may lead to inefficient or unsafe handling."
		},
		"Suggestions": {
			"Description": "A list of suggested adjustments or improvements to the VLM output. Each suggestion should be accompanied by a confidence level indicating the reliability of the recommendation.",
			"Example": [
			{
				"id": "A",
				"text": "Specify the exact gripping position on the headphone, such as the headband or ear cups, to ensure a secure hold.",
				"confidence": 0.9
			},
			{
				"id": "B",
				"text": "Define the coordinates or reference points for the gripping action to enhance precision.",
				"confidence": 0.8
			},
			{
				"id": "C",
				"text": "Include details about the gripping force and approach angle to prevent accidental drops or damage.",
				"confidence": 0.7
			}
			]
		},
		"Confidence": {
			"Description": "A numerical value representing the confidence in the output of SLM, based on the analysis of the VLM output and the task constraints.",
			"Example": 0.85
		},
		"Prompt for VLM": {
			"Description": "A command that can be fed into VLM to refine its output. This should include all the necessary details for VLM to adjust its instructions.",
			"Example": "Please specify the exact position on the headphone to grasp (e.g., the headband), and provide the coordinates or reference points for precise handling."
		}
	}
\end{lstlisting}

\subsection{VLM Output Example}
\label{VLM-Output}
\begin{lstlisting}[language=json, style=jsonstyle, caption={Detailed Example VLM Output}, label={lst:vlm_detailed_output}]
	"scene_description": {
		"objects": [
		{
			"name": "headphone",
			"position": [0.5, 0.3, 0.2],
			"properties": {
				"fragility": "high"
			}
		},
		{
			"name": "headphone_stand",
			"position": [0.4, 0.2, 0.1],
			"properties": {
				"material": "plastic",
				"stability": "unstable"
			}
		},
		{
			"name": "obstacle",
			"position": [0.6, 0.4, 0.3],
			"properties": {
				"size": "medium",
				"type": "box"
			}
		}]
	},
	"task_steps": [
	{
		"step_id": "1",
		"action": "moveTo(0.3, 0.2, 0.5)",
		"description": "Move the robotic arm to the initial position near the headphone stand."
	},
	{
		"step_id": "2",
		"action": "analyse()",
		"description": "Analyze the scene to identify the headphone on the stand and calculate its position."
	},
	{
		"step_id": "3",
		"action": "moveTo(0.35, 0.25, 0.45)",
		"description": "Move the robotic arm to the calculated position for grasping the headphone."
	},
	{
		"step_id": "4",
		"action": "grasp(15)",
		"description": "Grasp the headphone with a secure force of 15N to avoid damage."
	},
	{
		"step_id": "5",
		"action": "liftTo(0.4, 0.3, 0.5)",
		"description": "Lift the headphone off the stand to a safe position. (This function is undefined.)"
	},
	{
		"step_id": "6",
		"action": "moveTo(0.5, 0.4, 0.2)",
		"description": "Move the robotic arm to the position above the user's hand."
	},
	{
		"step_id": "7",
		"action": "release()",
		"description": "Release the headphone into the user's hand."
	}
	],
	"issues": [
	{
		"description": "The function 'liftTo' is not recognized as a valid robotic function.",
		"step_id": "5",
		"suggestion": "Replace 'liftTo' with 'moveTo' using appropriate coordinates."
	}
	],
	"flag": "complete"
\end{lstlisting}

\subsection{SLM Output Example}
\label{SLM-Output}
\begin{lstlisting}[language=json, style=jsonstyle, caption={Detailed Example SLM Output}, label={lst:slm_detailed_output}]
	{
		"Feedback": {
			"Description": "The VLM output contains issues in function validity and parameter safety. The 'liftTo' function in step 5 is undefined, and the grasp force in step 4 exceeds the safe threshold for fragile objects.",
			"Details": [
			{
				"step_id": "4",
				"issue": "The grasp force of 15N is above the recommended threshold for the headphone.",
				"recommendation": "Adjust the grasp force to 5N to ensure safety."
			},
			{
				"step_id": "5",
				"issue": "The function 'liftTo' is not recognized as a valid robotic function.",
				"recommendation": "Replace 'liftTo' with 'moveTo' using appropriate coordinates."
			}
			]
		},
		"Suggestions": [
		{
			"id": "A",
			"text": "Replace 'liftTo(0.4, 0.3, 0.5)' with 'moveTo(0.4, 0.3, 0.5)' in step 5 to ensure the action is executable.",
			"confidence": 0.9
		},
		{
			"id": "B",
			"text": "Reduce the grasp force in step 4 to 5N to avoid damaging the headphone.",
			"confidence": 0.85
		},
		{
			"id": "C",
			"text": "Include a validation step after the grasp action to confirm successful gripping before proceeding to lift.",
			"confidence": 0.8
		}
		],
		"Confidence": {
			"Description": "A confidence score based on the reliability of the suggestions provided.",
			"Value": 0.85
		},
		"Prompt for VLM": {
			"Description": "Refine the task plan by addressing the identified issues and suggestions.",
			"Command": "Please correct the undefined function 'liftTo' in step 5, adjust the grasp force in step 4 to 5N, and add a validation step after the grasp action to confirm successful gripping."
		}
	}
	
\end{lstlisting}

\bibliographystyle{IEEEtran} 
\bibliography{references}

\newpage

\end{document}